\documentclass[letterpaper,10 pt,conference]{ieeeconf}  

\IEEEoverridecommandlockouts                              

\usepackage{graphicx}
\usepackage{amsmath}
\usepackage{amssymb}
\usepackage{bm}
\usepackage{hyperref}
\usepackage[capitalize,noabbrev]{cleveref}
\usepackage{booktabs}
\usepackage{multirow}
\usepackage{caption}
\usepackage{subcaption}
\usepackage{xcolor}
\usepackage{amsmath} 
\usepackage{algorithm}
\usepackage[noend]{algpseudocode}

\newtheorem{proposition}{Proposition}
\newenvironment{propositionwithtitle}[1]{%
  \begin{proposition}%
  \textbf{#1} \par%
}{%
  \end{proposition}%
}

\newcommand{\separ}{SePar}

\title{\LARGE \bf
Sequence Pathfinder for Multi-Agent Pickup and Delivery \\ in the Warehouse
}

\author{Zeyuan Zhao$^{1}$, Chaoran Li$^{1}$, Shao Zhang$^{1}$ and Ying Wen$^{1,*}$
\thanks{$^{1}$Shanghai Jiao Tong University
        }%
\thanks{*Correspondence Author, {\tt\small ying.wen@sjtu.edu.cn}}
}

\begin{document}

\maketitle
\thispagestyle{empty}
\pagestyle{empty}

\footnotetext{Appendix: \nolinkurl{anonymous.4open.science/r/ICRA2026-SePar/Appendix.pdf}}
\begin{abstract}

Multi-Agent Pickup and Delivery (MAPD) is a challenging extension of Multi-Agent Path Finding (MAPF), where agents are required to sequentially complete tasks with fixed-location pickup and delivery demands. Although learning-based methods have made progress in MAPD, they often perform poorly in warehouse-like environments with narrow pathways and long corridors when relying only on local observations for distributed decision-making. Communication learning can alleviate the lack of global information but introduce high computational complexity due to point-to-point communication. To address this challenge, we formulate MAPF as a sequence modeling problem and prove that path-finding policies under sequence modeling possess order-invariant optimality, ensuring its effectiveness in MAPD. Building on this, we propose the Sequential Pathfinder (\separ{}), which leverages the Transformer paradigm to achieve implicit information exchange, reducing decision-making complexity from exponential to linear while maintaining efficiency and global awareness. Experiments demonstrate that \separ{} consistently outperforms existing learning-based methods across various MAPF tasks and their variants, and generalizes well to unseen environments. Furthermore, we highlight the necessity of integrating imitation learning in complex maps like warehouses.
\end{abstract}

\section{Introduction}

As an NP-hard problem, Multi-Agent Path Finding (MAPF) \cite{van2011reciprocal, cui2012pareto} has many real-life applications, such as search \& rescue \cite{baxter2007multi, berger2015innovative} and surveillance and warehouses \cite{nagorny2012service}. 
The goal of the naive one-shot MAPF is to generate collision-free paths for a set of agents from initial positions to specified goals on a given graph (usually a grid world), with the aim of minimizing a well-defined objective such as makespan (i.e., the time until all robots are on target). 
In MAPF, when agents are continuously assigned new target positions and are required to sequentially compute paths to reach these positions, the problem is more complex and defined as Lifelong Multi-Agent Path Finding (LMAPF) \cite{ma2017lifelong, liu2019task}.
A particularly demanding instance of LMAPF is Multi-Agent Pickup and Delivery (MAPD) \cite{ma2017lifelong, liu2019task}, requiring agents to reach designated locations and collaborate with humans for pick-up and drop-off tasks. 
Consequently, the key challenge in MAPD is completing a single item delivery involves at least two uninterrupted path-finding processes, which lead to an increased difficulty in finding solutions.

\begin{figure}[t]
\centering
\includegraphics[width=0.85\columnwidth]{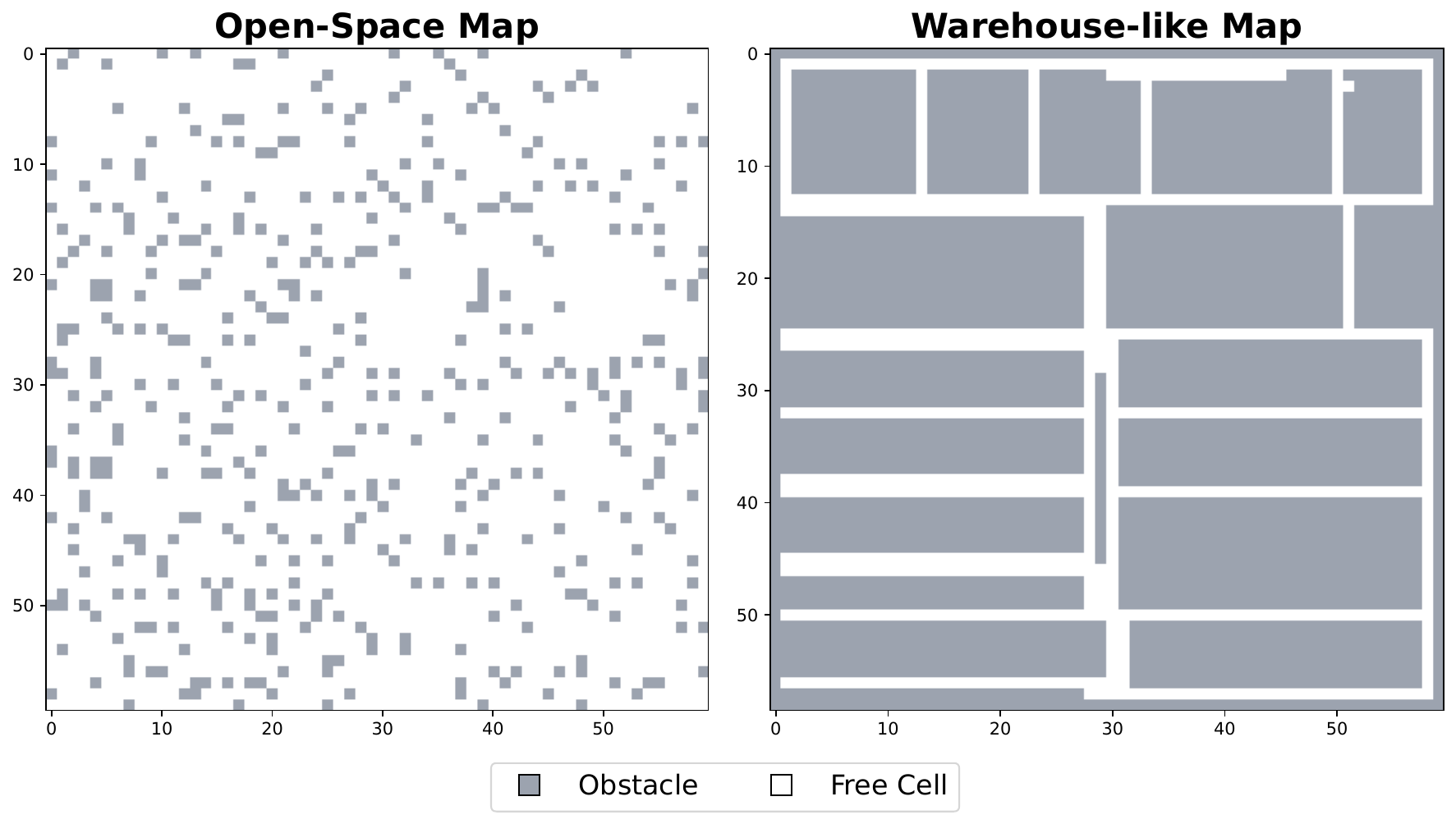} 
\caption{Comparison of two grid environments. 
Left: an open-space map with wide free space and multiple routing options. Right: a warehouse-like map with long, corridors and narrow pathways. These contrasting structures illustrate the intuitive difference in pathfinding difficulty.
}
\label{compare}
\vspace{-15pt}
\end{figure}

Although learning-based methods achieve recognized success in MAPD \cite{lau2022multi}, they still suffer from the complexity introduced by the narrow pathways and long corridors, which is very common in real world application represented by the warehouse.
As shown in \Cref{compare}, the map on the right is a typical warehouse layout, in sharp contrast to the map on the left with wide free space. Such warehouse-like environments are intuitively more challenging for MAPF, as agents must frequently traverse single-file corridors. 
However, current learning-based methods trained the agents only relying on local observations without taking global information into account face a risk of conflicts due to such a distributed execution.
For instance, PRIMAL2 \cite{damani2021primal} incorporate partial information from other agents within the Field of View (FOV) into the observation space, but still fail to achieve satisfactory performance. 
Furthermore, some studies have designed explicit communication modules to enhance information exchange among agents to reduce conflicts for better coordination \cite{ma2021distributed, ma2021learning, li2022multi}.
While these modules effectively reduce collisions, the reliance on complete or partial point-to-point communication incurs additional computational overhead, which reduces the learning efficiency of models.

To fully leverage global information while maintaining learning efficiency, a natural approach is to abstract the learnable MAPF problem as a sequence modeling task. 
Sequence modeling employs an autoregressive policy, enabling each agent to access the actions of its preceding agents and make optimal decisions based on this information \cite{wen2022multi}. 
In this way, the complexity of multi-agent reinforcement learning (MARL) is reduced from multiplicative to additive, resulting in linear complexity \cite{wen2022multi}. 
However, MAPD is highly sensitive to the decision order of agents \cite{ma2019searching, zhang2022learning}; we ensure the validity of using a sequence model for MAPD by \textbf{proving the order-invariant optimality of autoregressive path-finding policies}—i.e., any permutation of the order of agents yields the same joint optimal action.
Then we address the MAPF problem by introducing the \textbf{Sequence Pathfinder (\separ{})}, a Transformer-based pathfinding framework that generates the joint action of agents by mapping their observation sequences to action sequences. The mapping involves encoding the observation sequences into refined observation sequences that capture high-level interactions among agents via the attention mechanism, as well as accessing the actions of predecessors. Through these operations, implicit information exchange among agents is achieved. \separ{} is trained using a mixed approach that simultaneously learns from environmental feedback and imitates expert policies.

We compared our approach with various learnable methods and planning-based algorithms. Training and evaluation were conducted in two environments: one is our warehouse simulator, which is from real-world scenarios, and the other is POGEMA \cite{skrynnik2024pogema}, which integrates the well-known MAPF benchmark from MovingAI. 
The evaluation results indicate that: i) In both the MAPD tasks within our warehouse simulator and the MAPF and LMAPF tasks in POGEMA, \separ{} outperforms the vast majority of other learning-based baselines. 
This advantage becomes particularly pronounced when the number of agents reaches 256 or more, where the performance of competing learning-based methods drops to only 7\% - 15\% of that of \separ{}.
ii) Imitation learning is essential for training MAPF solvers in highly structured environments with narrow pathways and long corridors; iii) Compared to other learnable baselines, \separ{} demonstrates notable generalization capabilities on unseen maps.

\section{Related Work}

Learning-based MAPF methods are generally applicable to both one-shot and lifelong tasks, without making a strict distinction between the two.
Representative works, PRIMAL \cite{sartoretti2019primal} and PRIMAL2 \cite{damani2021primal}, use a decentralized policy trained through reinforcement and imitation learning.
All agents share a network, with reinforcement learning (RL) trained through A3C \cite{mnih2016asynchronous} and imitation learning (IL) using ODrM* \cite{ferner2013odrm} as the expert for behavioral cloning. However, these distributed policies do not inherently endow the agents with mechanisms for direct collaboration, relying only on local observations for decision making. As a result, it often leads to inefficient behaviors in warehouse environments with narrow pathways and long corridors.

Another type of learning-based methods trains decentralized policies from scratch using RL without IL. For instance, Follower \cite{skrynnik2024learn} assumes no reliable inter-agent communication and thus avoids using centralized expert policies, instead embedding an improved single-agent A* algorithm into each agent’s observations. Although it adapts to unknown maps, this real-time planning is computationally expensive. Moreover, in complex, highly structured maps requiring strong coordination, such as warehouses, distributed methods like Follower are unable to exploit global information, thereby increasing the risk of conflict.

Because distributed policies inherently limit information exchange among agents, thereby hindering effective coordination, many studies have sought to enhance the solution quality by incorporating communication learning when available.
Distributed Heuristic Learning with Communication (DHC) \cite{ma2021distributed} employs graph convolution for agent communication but restricts neighbor count, reducing efficiency in complex settings.
Decision Causal Communication (DCC) \cite{ma2021learning} enables selective, request-response communication based on local observations, supporting decentralized execution but limited by FOV.
PrIoritized COmmunication Learning (PICO) \cite{li2022multi} combines planning with priority prediction via ODrM* imitation and ad-hoc routing to reduce collisions, though scalability remains an issue.
SCRIMP \cite{wang2023scrimp} leverages a modified Transformer as the communication block to share messages, enhancing cooperation and scalability at the cost of higher computation.
Although these communication-learning approaches have shown notable gains in performance, their reliance on point-to-point message passing often introduces considerable computational overhead, in turn leaving scalability as a key limitation.

\section{Problem Formulation}
\label{problem_formulation}
In this section, we present how the MAPF problem is cast into a MARL framework. We detail the observation and action spaces of agents, the reward structure and the neural network that represents the policy to be learned.
\begin{figure*}[t]
\centering
\includegraphics[width=1.94\columnwidth]{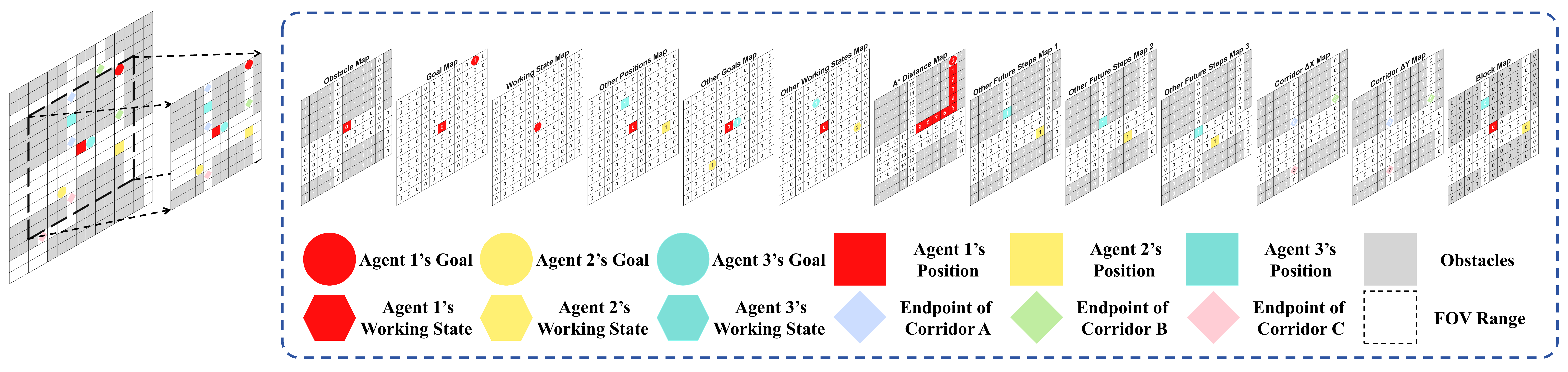} 
\caption{Observation space of the agents (here for agent 1, in red). 
Agents and corridor endpoints are represented by different colors. Circles, squares, and hexagons denote agents’ goals, positions, and statuses, while diamonds mark corridor endpoints.
}
\label{obs}
\vspace{-15pt}
\end{figure*}

\vspace{-5pt}

\subsection{LMAPF as a Dec-POMDP}
The traditional MAPF problem is represented by a tuple $<\mathcal{G},\mathcal{U},\mathcal{T}>$, where $\mathcal{G}=<V,E>$ is an undirected graph representing a workspace of $n$ agents. $\mathcal{U} = \{u^1, \cdots , u^n\} \subset V$ and $\mathcal{T} = \{\tau^1, \cdots, \tau^n\} \subset V$ contain $n$ unique start vertices and $n$ unique target vertices, one for each agent. The timeline is discretized into timesteps. At each time step, the agents either stand still or move to a neighboring vertex. The solution to the problem is a collision-free, shortest joint path from $\mathcal{S}$ to $\mathcal{T}$, consisting of $n$ single-agent paths. Through a series of feasible actions, agents arrive at their goal locations. Lifelong MAPF is a variant of MAPF in which, upon reaching a goal, an agent is immediately assigned a new goal and continues its path finding operation. Therefore, we use $\tau^i_t$ to denote the target vertex of agent $i$ at timestep $t$. Since our work focuses solely on path finding, the MAPD discussed in this paper is in fact the more complex LMAPF. Agents continuously receive new orders, and the delivery of each order is treated as a two-stage path finding process: from the current position to the pickup point, and then from the pickup point to the drop-off point.

This problem is framed as a Decentralized Partially Observable Markov Decision Process (Dec-POMDP) \cite{bernstein2002complexity} represented by the tuple $M = ⟨S, A, \mathbb{N}, P, R, O, \gamma⟩$. At each time step, each agent $i \in \mathbb{N}$, where $\mathbb{N} = \{1, \cdots , n\}$, observes an observation $o^i\in O$ from the global state $s$, forming the joint observation $\bm{o}=(o^1,\cdots,o^n)\in O^n$. The agent then chooses an action $a^i \in A$ via its policy $\pi^i$, the $i$-th component of the joint policy $\bm{\pi}$, yielding a joint action $\bm{a} \in A^n$. The transition function $P(s'|s, \bm{a}): S \times A^n \times S \rightarrow [0, 1]$ determines the (stochastic) variation of the environment driven by this joint action. The reward of each agent is determined by its reward function $R^i(\bm{o}, \bm{a}) : O^n \times A^n \times \mathbb{N} \rightarrow \mathbb{R}$. The discount factor $0 \leq \gamma \leq 1$ determines the importance of the future reward for the current state.

\vspace{-5pt}

\subsection{RL Environment Setup}
We introduce the warehouse simulator, which is an RL environment abstracted from real-world scenarios. In this section, we introduce three key components of this environment: the observation space, action space, and reward structure. 
Additionally, to formally capture the differences in complexity between different maps and highlight the characteristics of warehouse-like maps with long corridors and narrow pathways, we define a new metric called the Path Finding Complexity Index.

\subsubsection{Observation Space}

The workspace of agents is a 2D grid of size $h\times w$ with obstacles marked as '1' and free cells as '0'. 
Each agent observes only within its FOV ($13\times m\times m$), and the observation space is adapted from PRIMAL2 and refined for MAPD.
The first six matrices encode nearby obstacles, the agent’s goal and working state (Picking, Dropping, Idle), and the positions, goals, and states of other agents.
The remaining seven capture the agent’s predicted path, the predicted positions of other agents for the next three steps, coordinate differences between corridor endpoints, and whether a corridor is blocked.
Figure \ref{obs} illustrates an FOV example.

\subsubsection{Action Space and Reward Structure}
The action space of the environment is $\{0,1,2,3,4\}$, where each number corresponds to a specific action: 0 represents NOOP, 1 is LEFT, 2 is RIGHT, 3 is UP, and 4 is DOWN. The reward structure is defined as follows: moving or staying results in a penalty of -0.3, collisions incur a penalty of -2, and reaching the goals provides different rewards depending on the action—either 5.0 for picking and 5.0 for dropping, or 0.0 for picking and 10.0 for dropping. 
The MAPD task comprises two phases: picking and dropping, leading our work to address two scenarios: the first treats these phases as distinct pathfinding tasks with separate rewards for reaching their targets, while the second awards the final reward only after completing the dropping phase. Additionally, to accommodate different requirements, our work considers multiple reward modes: GLOBAL, INDIVIDUAL and PARTIAL. GLOBAL indicates that all agents share a common reward function, INDIVIDUAL signifies that each agent has a distinct reward function, and PARTIAL lies between the two: $\tilde{R}^i_t=\alpha R^i_t+(1-\alpha)\sum_{j\in \mathbb{N}\backslash \{i\}}R^j_t$, where $\tilde{R}^i_t$ is the partial reward of agent $i$, $R^i_t$ is the individual reward received by agent $i$ from the environment, and $\alpha$ is a coefficient to regulate the weight of the individual reward.

\subsubsection{The Complexity of Maps}
The difficulty of the MAPF task is directly proportional to both the number of agents and the complexity of the map. 
However, most works do not adequately measure the complexity of the map, involving only two simple indicators: map size and obstacle density. To more accurately represent the complexity of a map, we propose a new metric called the \textbf{Path Finding Complexity Index (PFCI)}, which is a tuple consisting of the \textbf{effective edge sparsity} and corridor length:
\begin{equation}\label{pfci}
    \begin{aligned}
          PFCI =  (l_{corr}, v_{e}),
    \end{aligned}
\end{equation}

\begin{figure}[t]
\centering
\includegraphics[width=0.8\columnwidth]{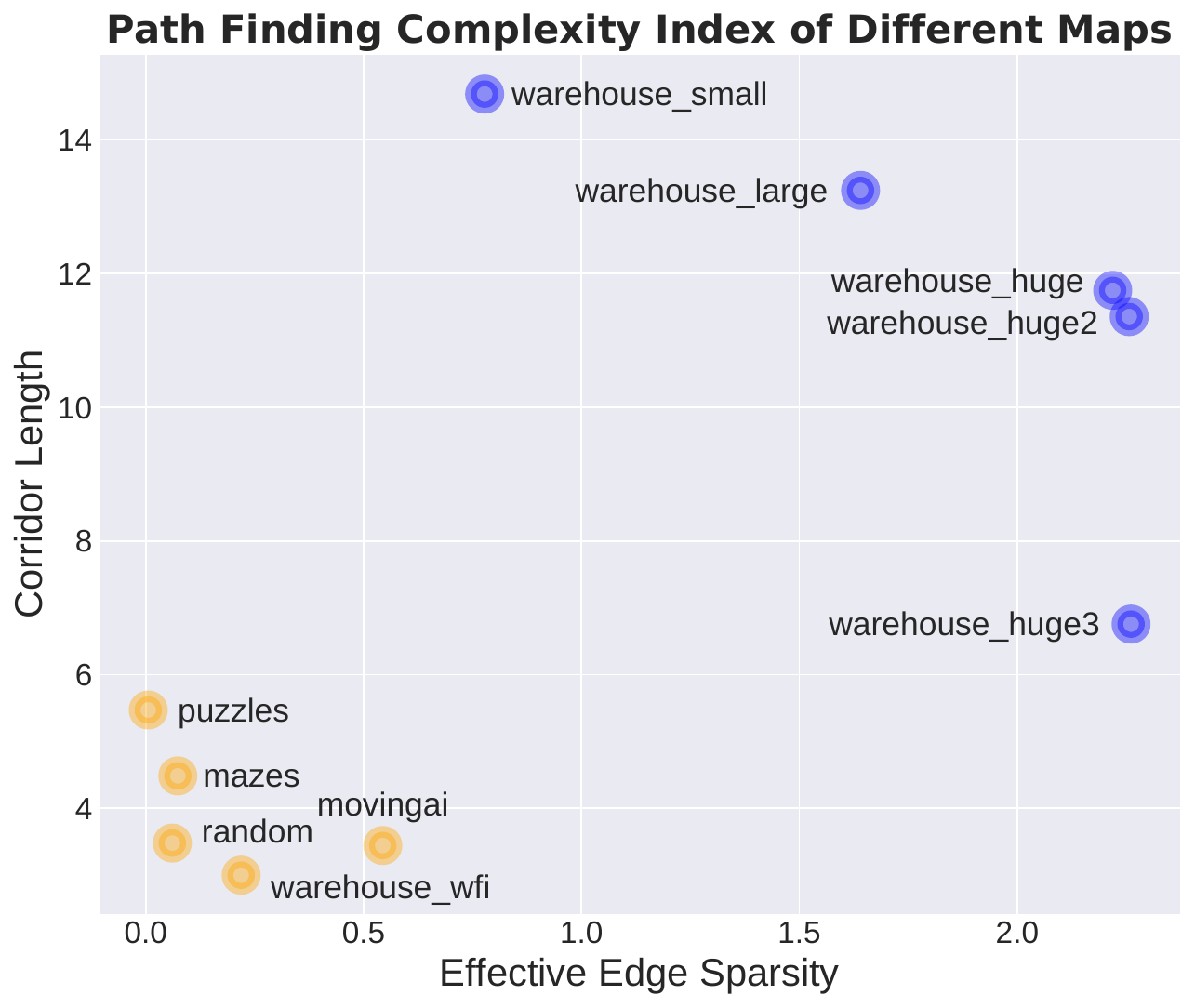} 
\caption{Path Finding Complexity Index of maps.
The orange ones are from the MAPF benchmark named POGEMA, and the blue ones are from our warehouse simulator.
}
\label{pfcis}
\vspace{-15pt}
\end{figure}

where $v_{e} = \alpha \times (\rho_{e} \times \rho_{t})^{-1}$, and $\alpha$ is the coefficient, $l_{corr}$ is the corridor length and $v_{e}$ is the effective edge sparsity. The effective edge sparsity is used to measure the narrowness of pathways. Specifically, $v_{e}$ is the inverse of the product of $\rho_{e}$ and $\rho_{t}$, $\rho_{t}=1-\rho_{o}$ represents the traversable area density, and $\rho{o}$ represents the obstacle density. We denote the number of nodes within the traversable area as $|V|$, and the number of edges connected by these nodes is $|E|$. Then the edge density is given by $\rho_{e}=\frac{2|E|}{|V|(|V|-1)}$. The PFCI different maps are shown in \Cref{pfcis}. The maps involved in it are partly from the well-known MovingAI benchmark in the MAPF community, which are integrated into POGEMA, and partly from our warehouse simulator. All map data are detailed in \Cref{metrics} of \Cref{app:pfci}. The maps are sourced from POGEMA and our warehouse simulator, respectively.

\section{Learning Methods}
\label{methods}
\subsection{MAPF as a Sequence Modeling Problem}
With the emergence of powerful architectures such as Transformers \cite{vaswani2017attention}, the significance of sequence modeling techniques has garnered increased attention within the reinforcement learning (RL) community \cite{chen2021decision, wen2022multi}. In this work, we approach the RL-based MAPF problem as a sequential decision-making task. The close relationship between MARL and sequential decision-making is fundamentally rooted in the multi-agent advantage decomposition theorem \cite{kuba2021settling}.
This theorem reveals a crucial property: when each agent is aware of the actions of its predecessors, maximizing the local advantage function for each agent is equivalent to maximizing the joint advantage. This implies that given a sequence of agents, each agent can access the actions of its predecessors and make optimal decisions based on this information. This establishes a multi-agent sequential decision-making paradigm, which notably reduces the complexity of the MARL problem from multiplicative to additive, resulting in linear complexity. \cite{wen2022multi}
A natural insight is that the decision order of agents within a sequence model may influence the upper bound of the performance of policies. Motivated by this intuition, we formally propose the following proposition regarding the impact of decision order.

\begin{propositionwithtitle}{Order-invariant Optimality of Autoregressive Pathfinding Policies}
\label{Order-invariant}
Let $n\ge 1$ agents act simultaneously in the environment. For any permutation $\sigma\in S_n$, an autoregressive policy can be defined as:
\begin{equation}
    \begin{aligned}
        \pi^{\sigma}_\theta(a^{1:n}\mid \bm{o})\stackrel{\text{def}}=\prod_{k=1}^{n}\,\pi_\theta\big(a^{\sigma[k]}\mid \bm{o}, a^{\sigma[1]},\ldots,a^{\sigma[k-1]}\big),
    \end{aligned}
\end{equation}

where $S_n$ is the set of all possible permutations of $n$ elements, $\sigma[k]$ represents the $k$-th element of permutation $\sigma$ and $\theta$ is the policy parameter. Then
\begin{equation}
    \begin{aligned}
        \forall \bm{o}\in O, \forall \sigma,\nu\in S_n, \implies  \pi^{\sigma}_{\theta}(\cdot\mid \bm{o})=\pi^{\nu}_{\theta}(\cdot\mid \bm{o}).
    \end{aligned}
\end{equation}
    
Let $f(\cdot)$ denote any objective function that depends only on the induced autoregressive pathfinding policy, then

\begin{equation}
    \begin{aligned}
        \forall \sigma,\nu\in S_n\implies\sup_{\theta} f\big(\pi^{\sigma}_{\theta}\big)=\sup_{\theta} f\big(\pi^{\nu}_{\theta}\big).
    \end{aligned}
\end{equation}

Consequently, the optimal makespan in oneshot MAPF or throughput in lifelong MAPF by autoregressive policies is \textbf{independent of the decision order} $\sigma$.
\end{propositionwithtitle}

For proof see \Cref{app:proof_order}. 
Then to implement the decision-making paradigm, we leverage a classic sequence modeling technique: the Transformer. Specifically, we introduce TRansFormer Pathfinder (\separ{}), which is based on the Multi-Agent Transformer architecture. \separ{} generates the joint action for agents by mapping their observation sequences to action sequences.
This method first employs the attention mechanism to encode the observation sequences into refined observation sequences that capture high-level interactions among agents. Subsequently, it accesses and encodes the actions of predecessors to generate the current agent's action. This process facilitates efficient implicit information exchange among agents. 

In contrast to \separ{}, most other learnable methods execute in a distributed manner, allowing agents to make decisions based only on their local observations \cite{sartoretti2019primal, damani2021primal, skrynnik2024learn}. This limitation limits their ability to effectively utilize global information for high-quality coordination, making them ill-suited for real-world scenarios that are highly structured, such as warehouses with long corridors and narrow pathways. Approaches like DHC \cite{ma2021distributed}, DCC \cite{ma2021learning}, PICO \cite{li2022multi}, and SCRIMP \cite{wang2023scrimp} employ explicit communication modules to enhance collaboration quality; however, as the number of agents increases, the computational overhead of point-to-point communication among agents escalates significantly. In contrast, \separ{} maintains a linear complexity in its communication overhead, ensuring efficient model training.
\vspace{-5pt}

\subsection{Network Structure}
\begin{figure*}[t]
\centering
\includegraphics[width=1.76\columnwidth]{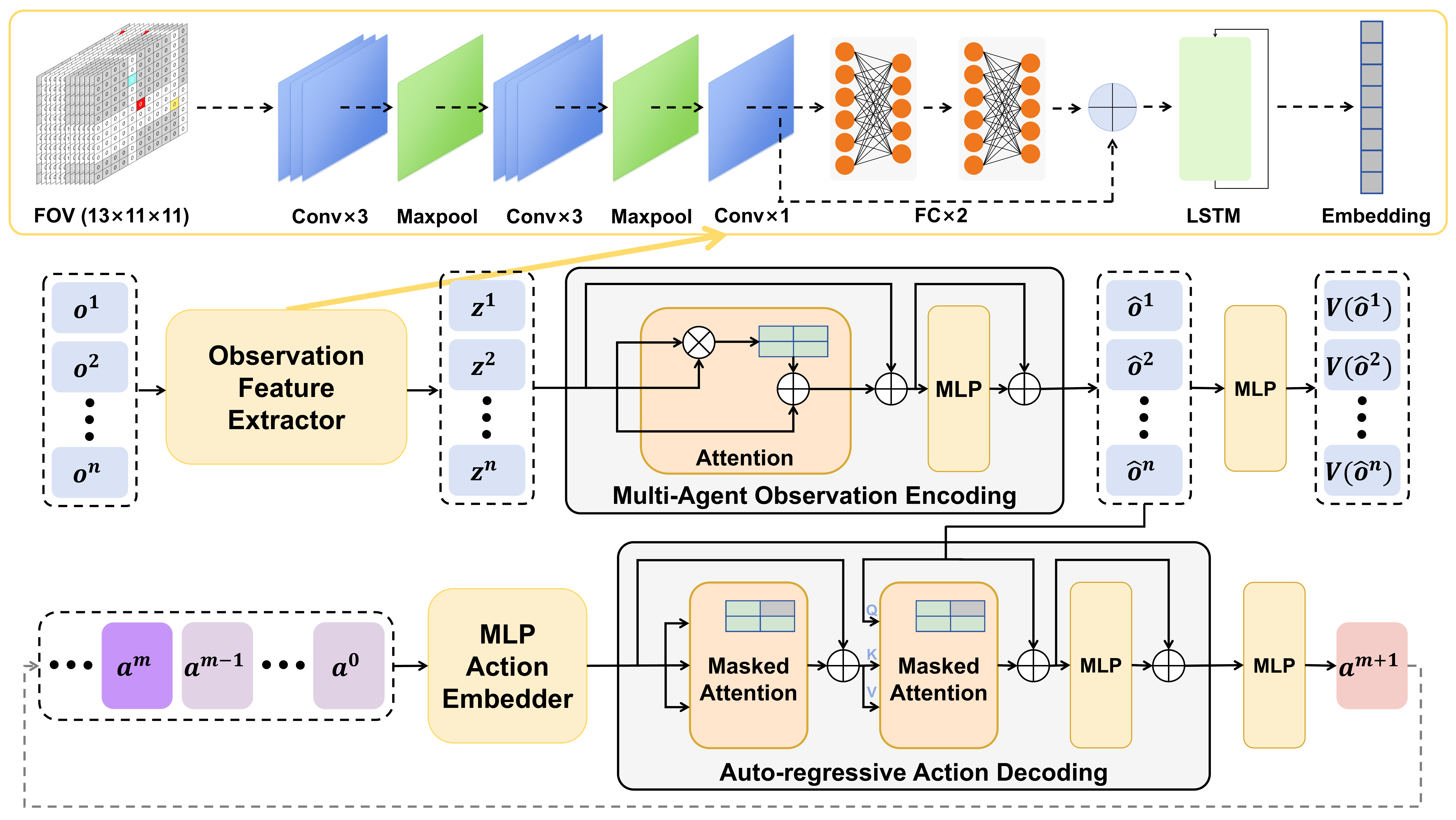}
\caption{Network Structure of the Transformer Pathfinder. 
At each step, the Observation Feature Extractor generates embeddings from agents' observations. The encoder refines these into new observations for the decoder, which uses masked attention to block access to subsequent agents' actions, ensuring each agent acts sequentially based on preceding agents.
}
\label{net}
\vspace{-15pt}
\end{figure*}

Figure \ref{net} shows the network structure of \separ{}, which consists of two modules: the observation feature extractor, and Multi-Agent Transformer.

\subsubsection{Observation Feature Extractor.}
It consists of two VGG blocks—each of which contains three convolutional layers and one max-pooling layer—along with one additional convolutional layer, two fully-connected layers, and one LSTM unit.At time step $t$, any agent $i$ receives the observation $o^i_t$ from the environment, which is then sequentially processed through two VGG blocks, and finally through a convolutional layer, to obtain the hidden information representation $x_0$. The representation $z_0$ is further passed through two fully-connected layers to obtain the hidden information representation $x_1$. $x_0$ and $x_1$ are then summed and combined with the unit's output hidden states from the previous time step, $h_{t-1}$ and $c_{t-1}$, respectively, and fed into the LSTM unit. This process gives each agent the ability to take past information into account when making decisions. The LSTM unit outputs the hidden state $h_{t}$ and $c_{t}$ and observation embedding $z_t$ for the current step.

\subsubsection{Multi-Agent Transformer (MAT).} MAT \cite{wen2022multi} is a novel Multi-agent Reinforcement Learning (MARL) architecture designed to achieve efficient multi-agent collaboration and decision-making by transforming MARL problems into sequence modeling problems. The core idea of the Multi-Agent Transformer (MAT) is to leverage the powerful modeling capabilities of sequence models, particularly the Transformer architecture, to address the complexity of the multi-agent interaction and decision-making process. Specifically, the Transformer handles the sequence of agents rather than the time series. That is, the encoder of the Transformer takes the observations (or observation embeddings) of all agents, $o^0_t,\cdots ,o^{n-1}_t$, as input. The encoder consists of several computational blocks, each focused on the self-attention mechanism. Observations are encoded as $\hat{o}^0_t,\cdots ,\hat{o}^{n-1}_t$, which can be referred to as \textbf{refined observations}. These refined observations encode not only information about the agents themselves but also the high-level interactions between them. Thus, the encoder facilitates implicit communication between agents. Unlike the standard Transformer, the refined observations are fed into a Multilayer Perceptron (MLP), and the output of this MLP constitutes a set of approximated value functions for the RL training phase.

The Transformer's decoder generates the actions for each agent one by one in an autoregressive manner, ensuring that each agent's decisions take into account the actions of the previous agent. Specifically, the inputs to the decoder combine the refined observations and the previous action sequences. The key part of each decoding block in the decoder is the masked self-attention mechanism; masking ensures that attention is computed only between sequences of agents with an index less than $i$. The second masked self-attention function computes the attention between action heads and refined observations. The final MLP in the encoder takes as input the action representations output by the previous block to determine the distribution of actions for the current agent, i.e., the policy $\pi^i(a^i|\hat{o}^{0:n-1}, a^{0:i-1})$.

\vspace{-5pt}

\subsection{PPO-based MAT Learning}
Implicit information exchange between agents is attributed to the attention mechanism, which is central to the Multi-Agent Transformer. The encoding of observations and actions is based on embedded queries $(q^1, \cdots , q^n)$ and keys $(k^1, \cdots , k^n)$ multiplied by a weight matrix computed where each weight $w(q^r, k^j ) = <q^r, k^j>$. In the encoder, the interrelationships between the agents, that is, $\hat{o}^{1:n}$, are extracted by the full weight matrix, as the attention mechanism is not masked. In the decoder, on the other hand, the masked attention mechanism employs a triangular matrix to capture $a^{1:i}$. Here, the weight $w(q^r, k^j )$ is set to 0 when $r < j$.

The output of the MAT is divided into two parts: the approximated value functions $V(\cdot)$ from the encoder and the policies $\pi$ from the decoder. The model is trained using Proximal Policy Optimization (PPO) \cite{schulman2017proximal}. The loss functions for the two parts are given by \Cref{enc loss,dec loss}, where $\phi$ and $\theta$ denote the parameters of the encoder and decoder, respectively. 
$\hat{A}_t$ is the estimated joint advantage function and $r^i_t(\theta)$ in \Cref{ratio} is the ratio of the policy $\pi^i_{\theta}$ to be updated to the original policy $\pi^i_{\theta_{old}}$.
\begin{equation}
    \begin{aligned}\label{enc loss}
        L_{\mathrm{Encoder}}(\phi) = &\frac{1}{nT}\sum_{i=1}^{n}\sum_{t=0}^{T-1} \left[ R(\bm{o}_t, \bm{a}_t)  \right. \\
        &\left. + \gamma V_{\overline{\phi}}(\hat{o}^i_{t+1}) - V_{\phi}(\hat{o}^i_{t}) \right]^2,
    \end{aligned}
\end{equation}
\begin{equation}\label{dec loss}
    \begin{aligned}
        L_{\mathrm{Decoder}}(\theta) = &-\frac{1}{nT}\sum_{i=1}^{n}\sum_{t=0}^{T-1}\min\left(r^i_t(\theta)\hat{A}_t, \right. \\
        &\left. \mathrm{clip}(r^i_t(\theta), 1\pm\epsilon)\hat{A}_t\right),
    \end{aligned}
\end{equation}

\begin{equation}\label{ratio}
    \begin{aligned}
        r^i_t(\theta)=\frac{\pi^i_{\theta}(a^i_t|\hat{o}^{1:n}_t,\hat{a}^{1:i-1}_t)}{\pi^i_{\theta_{old}}(a^i_t|\hat{o}^{1:n}_t,\hat{a}^{1:i-1}_t)}.
    \end{aligned}
\end{equation}

\vspace{-5pt}

\subsection{Imitation Learning (IL)}
Relevant work in robotics has demonstrated that combining IL with RL ensures more efficient and stable training, and improves solution quality \cite{gao2018reinforcement, nair2018overcoming, sartoretti2019primal}. 
In our task scenario, utilizing demonstrations from expert policies, such as heuristic planning algorithms, plays a very important role. 
Starting RL from scratch is a feasible solution, but it is only effective in tasks with lower difficulty, such as smaller map sizes, lower obstacle densities, and fewer agents, and results in slower learning speeds. 
An effective way to accelerate learning is to combine on-policy RL and off-policy IL. 
The high-quality data provided by the heuristic algorithm enables agents to quickly identify critical state-action pairs within the vast exploration space. 
Meanwhile, RL's exploration mechanism encourages agents to explore areas surrounding these pairs, thereby further enhancing policy quality. In our work, IL achieves this by minimizing the behavioral cloning loss (\Cref{bc loss}), which allows the agents to closely mimic the expert policies.
\begin{equation}\label{bc loss}
    \begin{aligned}
          L_{\mathrm{bc}}(\phi, \theta)=-\frac{1}{nT}\sum_{i=1}^{n}\sum_{t=0}^{T-1}\left[\log P(a^{*,i}_t|\pi^i_{\theta},\hat{o}^{1:n}_t;\phi,\theta)\right],  
    \end{aligned}
\end{equation}
where $a^{*,i}_t$ is the expert action from the heuristic algorithm. 

The combination of RL and IL components naturally motivates a unified training routine in \separ{} with the full pseudocode in \Cref{app:pse-code}.

\section{Experiments}

To evaluate the performance of the proposed method \separ{}, we first introduce key metrics of the task. 
Based on this metric and task classification, we conduct experiments comparing the proposed method with state-of-the-art algorithms on different maps. 
Training and evaluation for the MAPD task were conducted in our warehouse simulation environment, while those for MAPF and LMAPF tasks were carried out in the POGEMA \cite{skrynnik2024pogema} environment. 
\Cref{snapshot} in the appendix shows our warehouse mockup and simulation environment.
Each experiment was evaluated five times by running five environments with different random seeds, ensuring distinct agent start positions.

\vspace{-5pt}

\subsection{Metrics}

\begin{figure*}[h]
    \centering
    \begin{subfigure}[t]{0.66\columnwidth}
        \includegraphics[width=\textwidth]{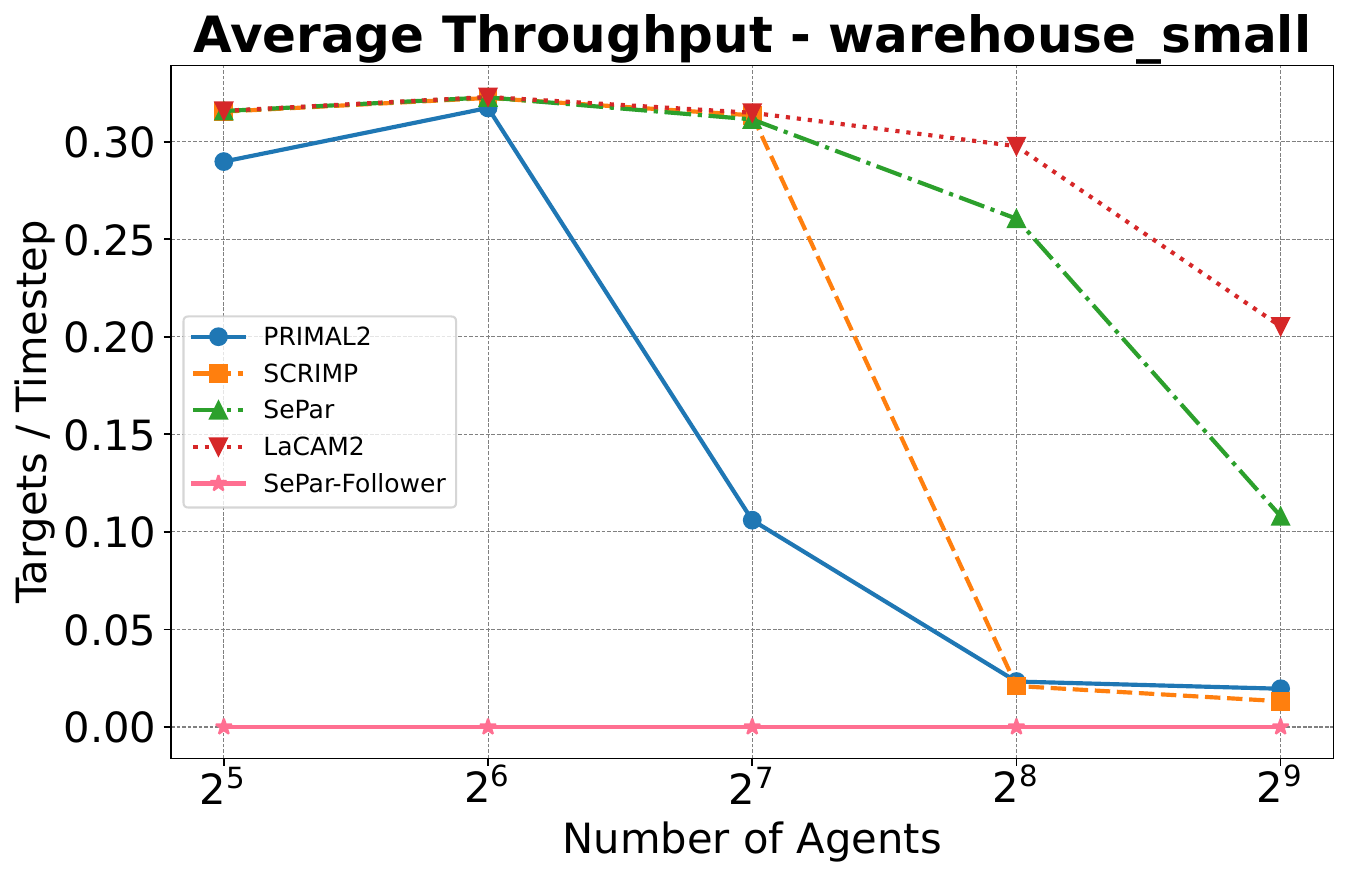} 
        \caption{}
        \label{wh_small}
        \end{subfigure}
    \hfill
    \begin{subfigure}[t]{0.66\columnwidth}
        \includegraphics[width=\textwidth]{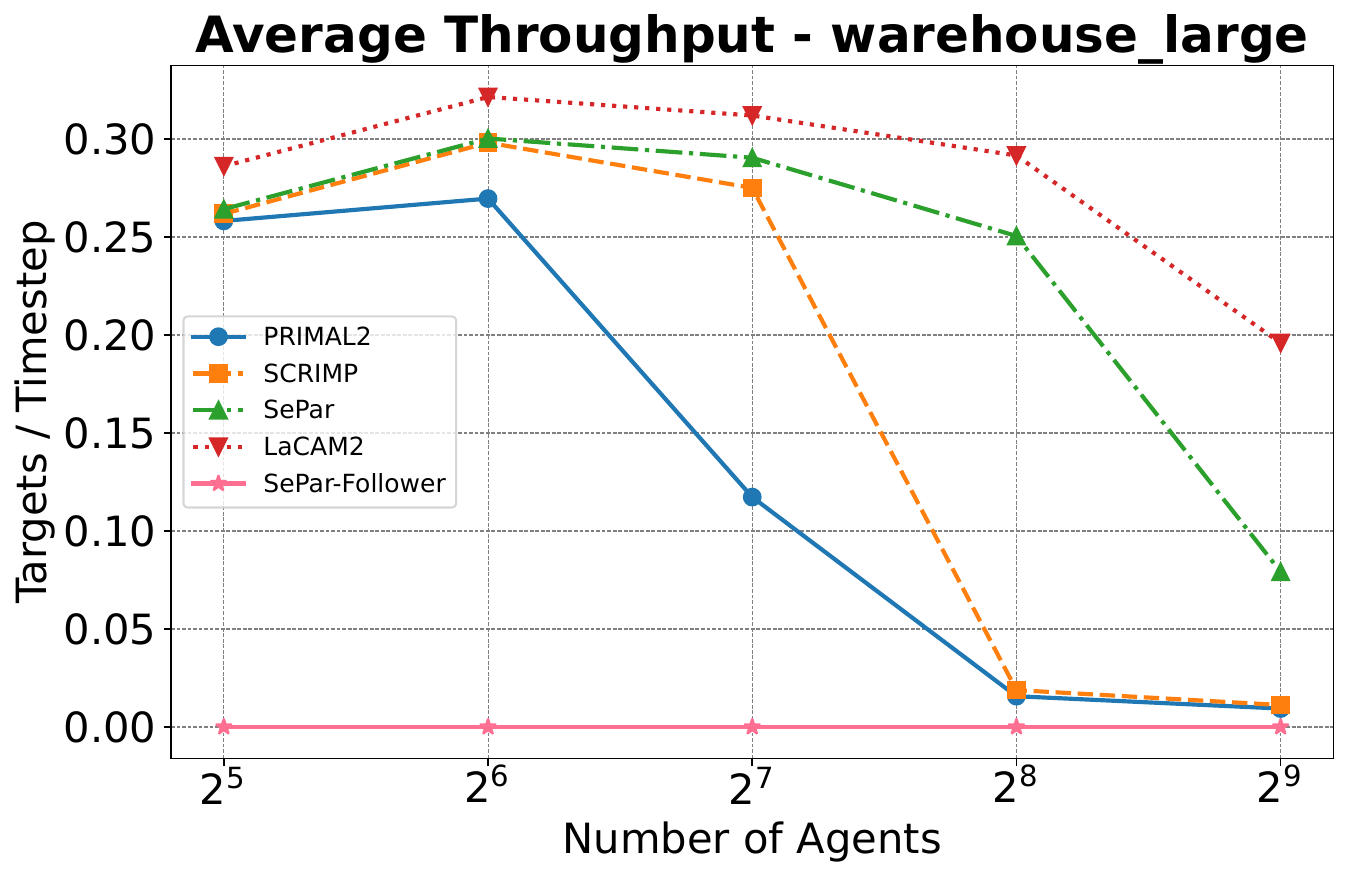} 
        \caption{}
        \label{wh_large}
    \end{subfigure}
    \hfill
    \begin{subfigure}[t]{0.66\columnwidth}
        \includegraphics[width=\textwidth]{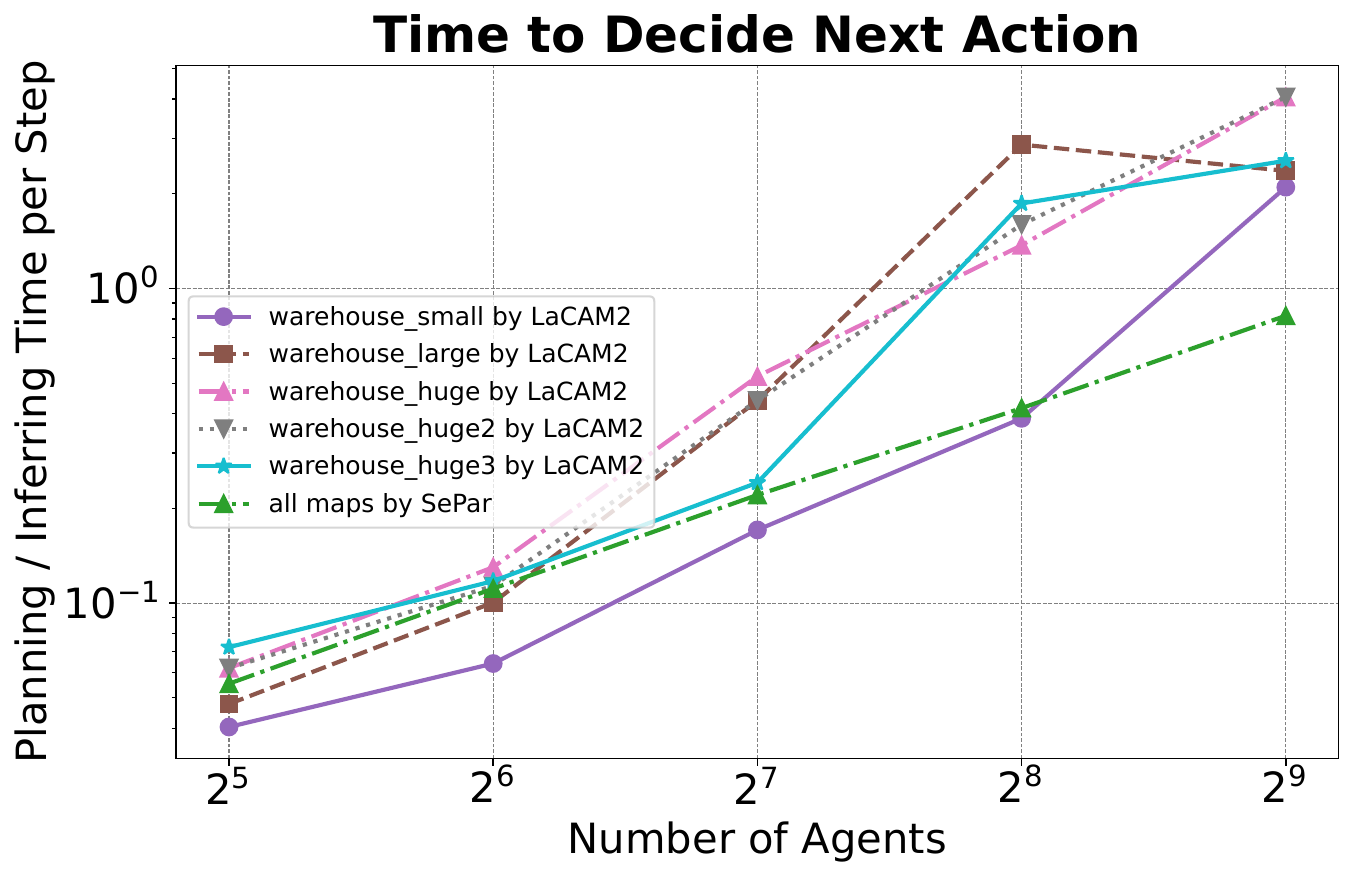} 
        \caption{}
        \label{runtime}
    \end{subfigure}
    \hfill

    \caption{Results in the warehouse simulation environment. 
    \Cref{wh_small} and \Cref{wh_large} present results on \textit{warehouse\_small} and \textit{warehouse\_large}. \separ{} outperforms the other learnable methods and shows scalability as agent numbers grow. \Cref{runtime} shows the average time per timestep for LaCAM2 and \separ{} to generate actions.
    }
    \label{mapd_results}
    \vspace{-7pt}
\end{figure*}

We evaluate algorithms on three primary metrics: \textbf{Success Rate}, \textbf{Sum of Cost (SoC)}, and \textbf{Throughput}.

For \textbf{one-shot MAPF} tasks, we focus on Success Rate and Sum of Cost (SoC), as agents are assigned a single goal each.
\begin{itemize}
    \item Success Rate measures the proportion of instances where the algorithm finds collision-free paths for all agents within the time limit.
    \item SoC calculates the total path length for all agents to reach their targets.
\end{itemize}

For \textbf{Lifelong MAPF} tasks, including \textbf{MAPD} tasks, we emphasize Throughput, defined as the average number of targets reached per unit time step \cite{damani2021primal}:

\vspace{-5pt}
\begin{equation}
    \begin{aligned}
          \text{Throughput} = \frac{\text{Total number of goals reached}}{\text{Total time steps}}.
    \end{aligned}
\end{equation}

In addition, we assess the \textbf{average runtime} of the algorithm at each time step.

\vspace{-5pt}

\subsection{MAPD Tasks in the Warehouse Simulator}
\label{results in wh}
\subsubsection{Experiment Setup}

\begin{figure*}[h]
    \centering
    \begin{subfigure}[t]{0.66\columnwidth}
        \includegraphics[width=\textwidth]{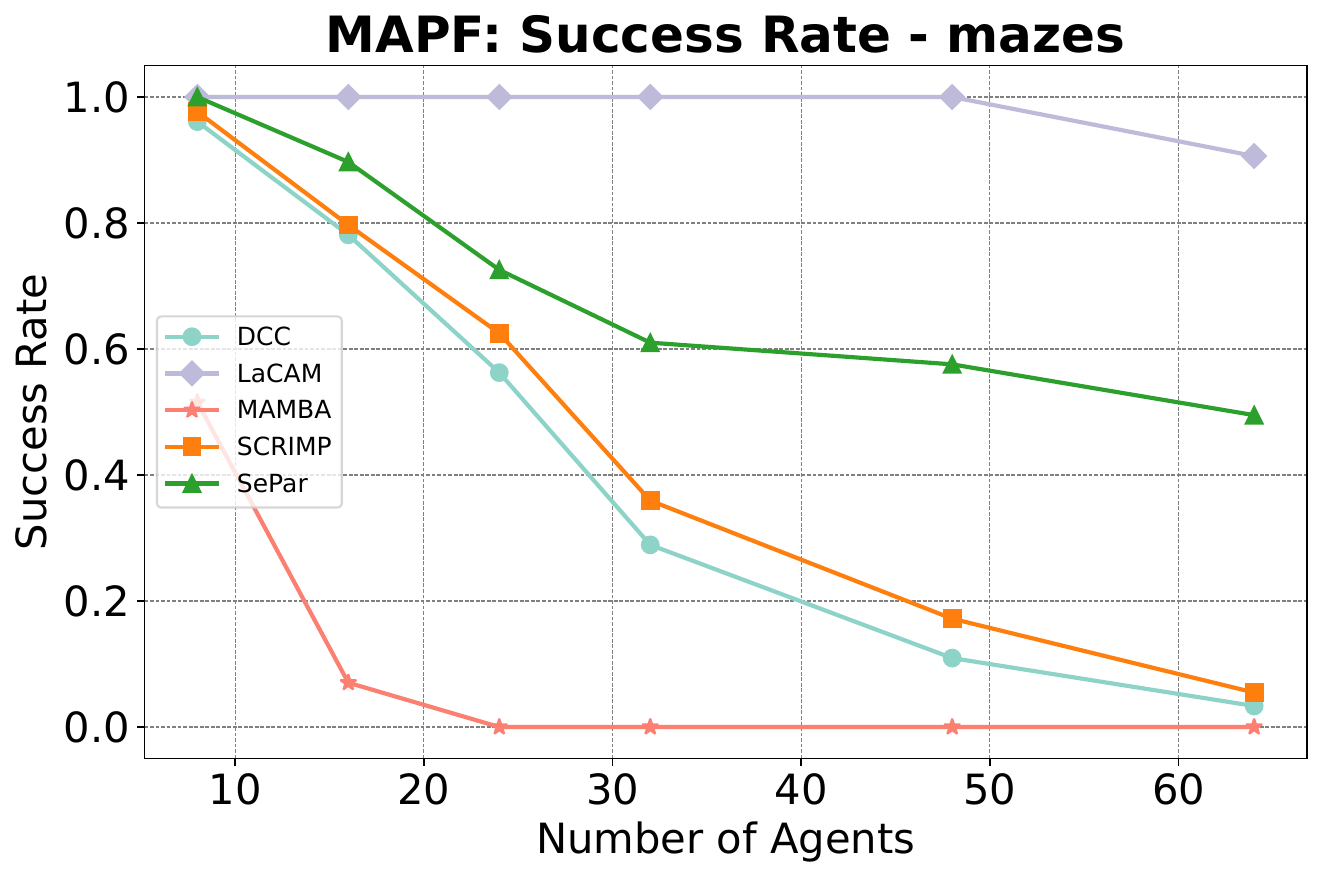}
        \caption{}
        \label{mazes_csr}
    \end{subfigure}
    \hfill
    \begin{subfigure}[t]{0.675\columnwidth}
        \includegraphics[width=\textwidth]{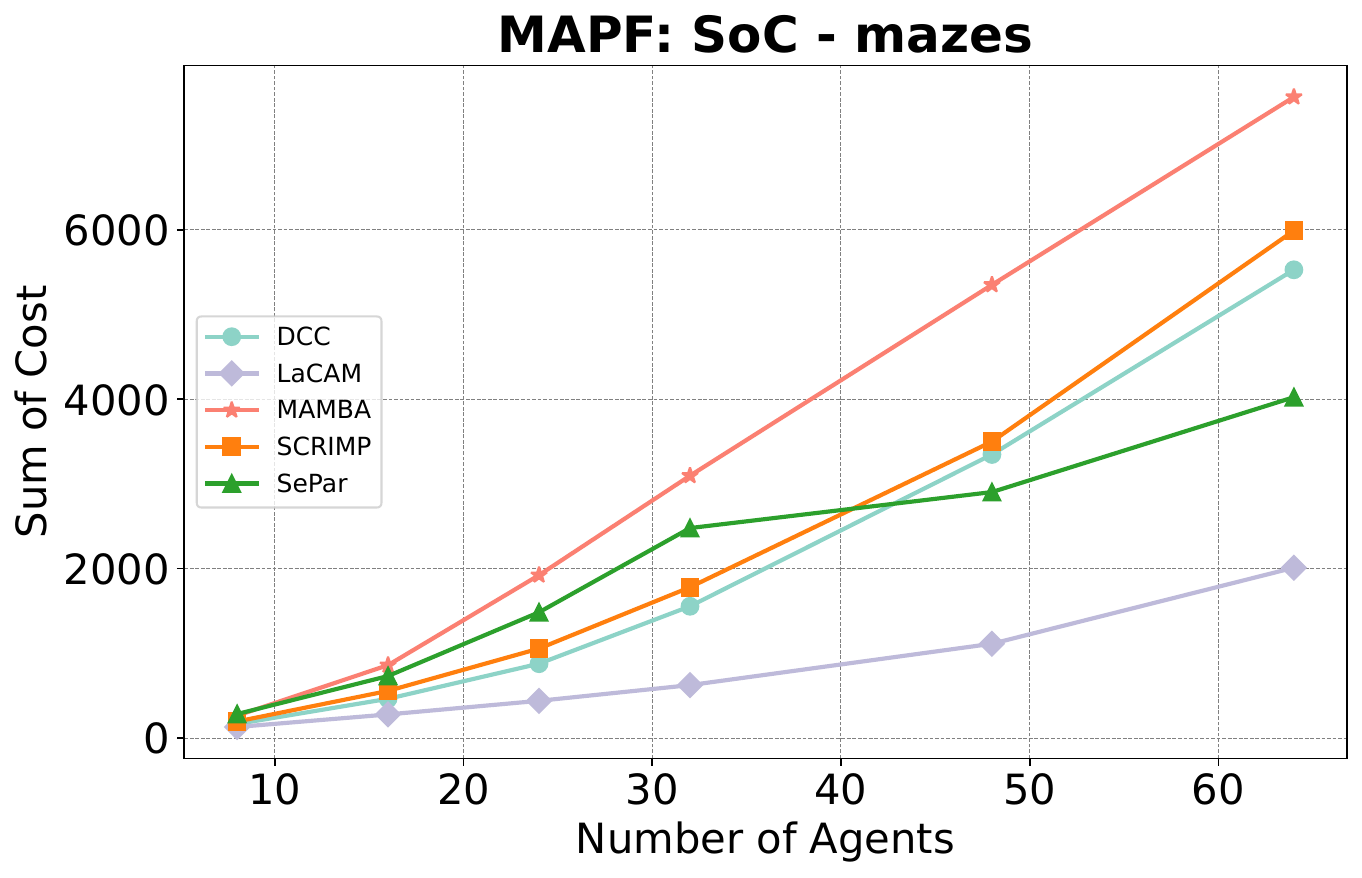}
        \caption{}
        \label{mazes_soc}
    \end{subfigure}
    \hfill
    \begin{subfigure}[t]{0.66\columnwidth}
        \includegraphics[width=\textwidth]{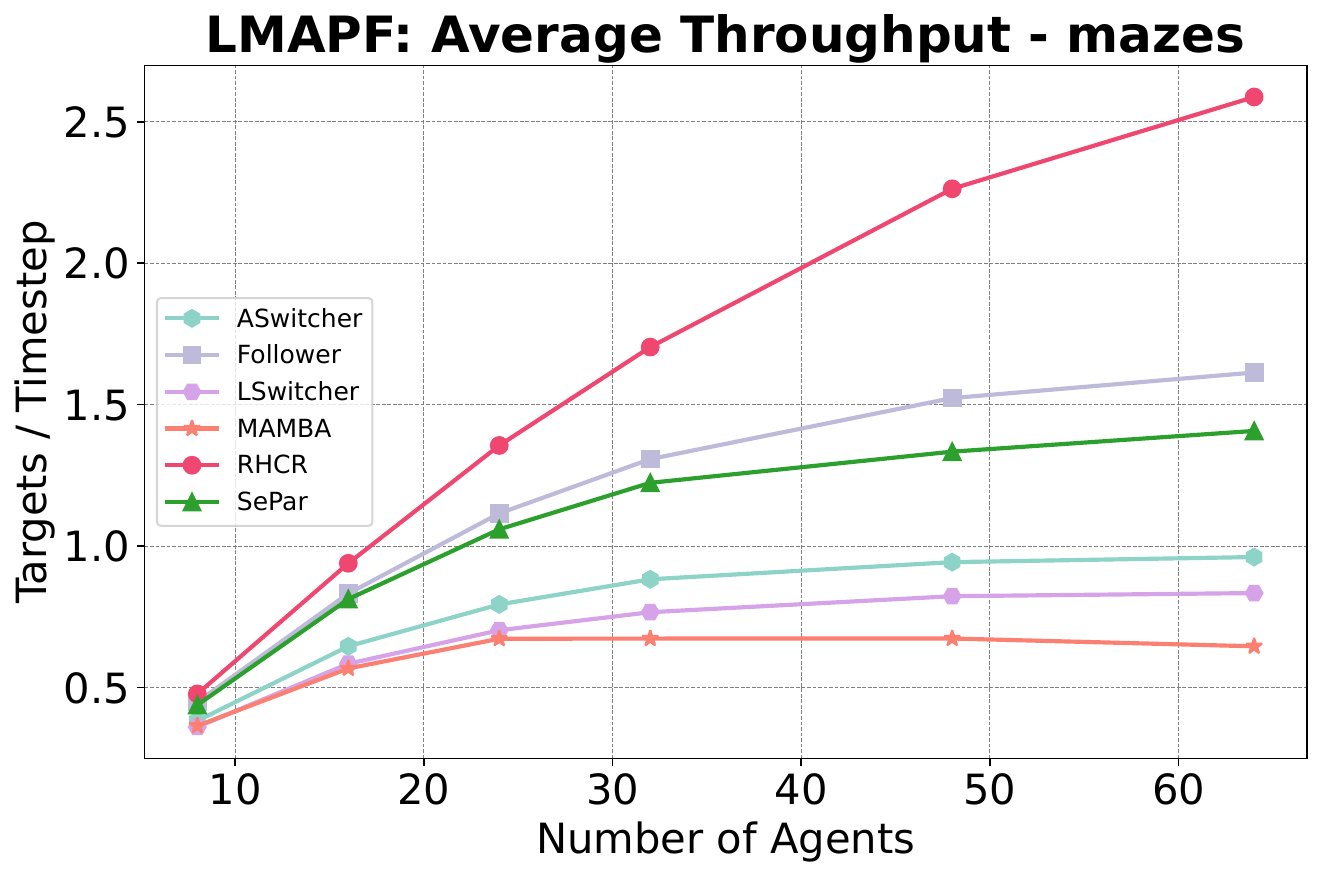} 
        \caption{}
        \label{mazes_ath}
    \end{subfigure}
    \caption{Results on \textit{mazes} from the POGEMA. 
    \Cref{mazes_csr,mazes_soc} shows MAPF results, and \Cref{mazes_ath} shows LMAPF results. In MAPF, \separ{} outperforms other learnable methods. In LMAPF, \separ{} ranks just behind Follower, a state-of-the-art learnable solver, and RHCR, a well-known planning solver.
    }
    \label{mapf_results}
    \vspace{-10pt}
\end{figure*}

In the experiments conducted in the warehouse environment, we compare \separ{} with state-of-the-art MAPF solvers — LACAM2 \cite{okumura2023improving}, SCRIMP, and PRIMAL2. LaCAM2 is a planning-based method that, although suboptimal, achieves a balance between success rate and planning time. The experiments are conducted on a system equipped with an AMD EPYC 7H12 64-core processor and an NVIDIA RTX 3090 GPU.
Notably, we also implement a version of \separ{} based on the concept of Follower, known as \separ{}-Follower. Specifically, we retain the network structure of \separ{} and the observation space of Follower, and also train the network from scratch without the use of imitation learning, much like Follower. The heuristic algorithm used to construct observations is LaCAM2, rather than the improved A* algorithm used by Follower, in order to enhance the effectiveness of the paths shown in the observations.

\begin{figure}[htbp]
\centering
\includegraphics[width=0.83\columnwidth]{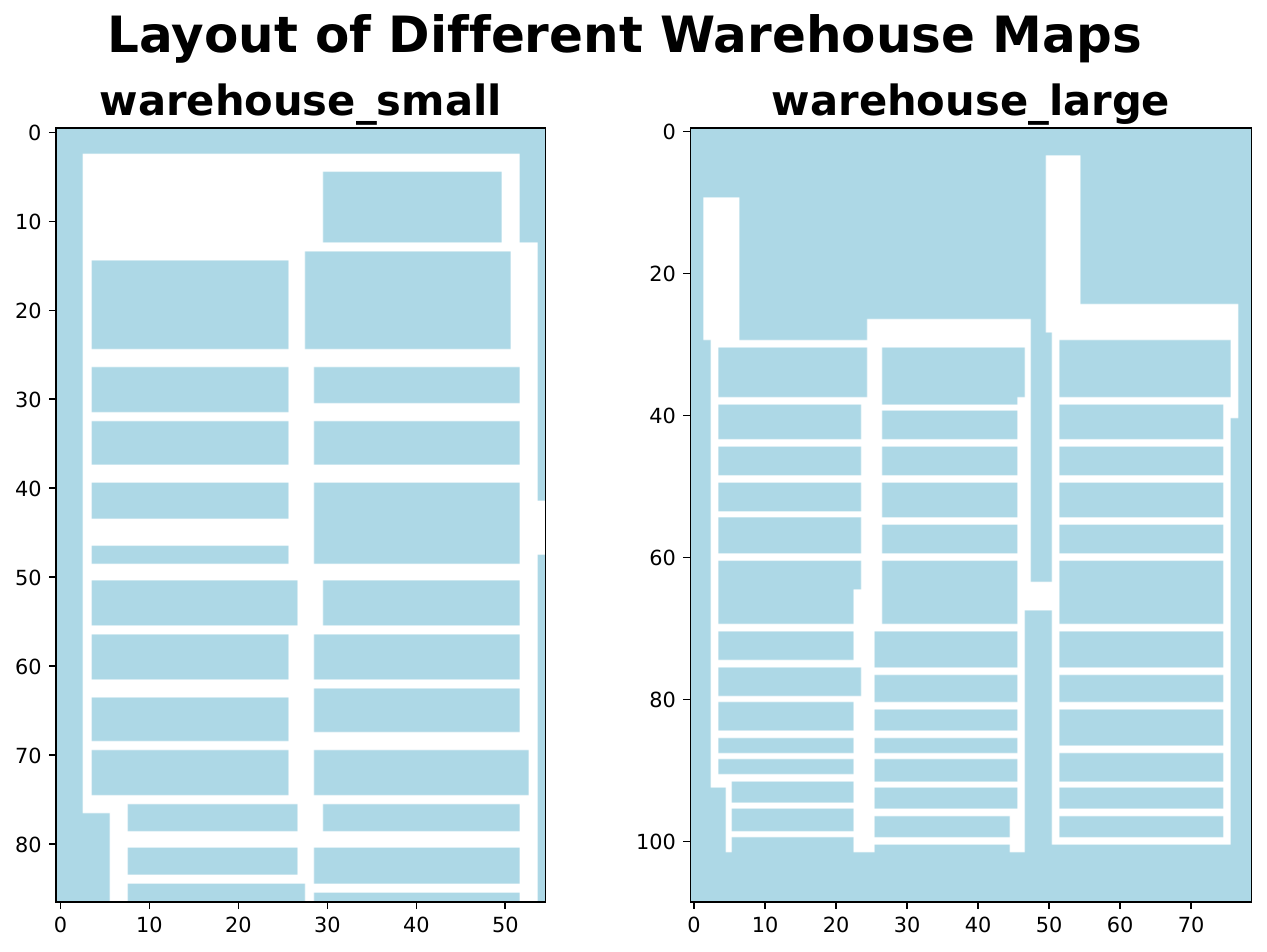}
\caption{Layout of different maps in the warehouse. The traversable areas are represented in white, while obstacles are indicated in light blue.}
\label{layout}
\vspace{-15pt}
\end{figure}

We trained all learnable methods on \textit{warehouse\_small} and \textit{warehouse\_large}, with their layouts depicted in \Cref{layout}. In these layouts, the traversable areas are represented in white, while obstacles are indicated in light blue. Both for training and evaluation, the starting positions of the agents are based on random generation, while the target positions are derived from the given order files. The number of agents involved in the tasks ranges from 32 to 512. The length of each episode during evaluation is set to 20,000.

In addition to normal performance, the generalizability of learnable algorithms is a matter of concern. The model is trained on the \textit{warehouse\_small} and \textit{warehouse\_large} and is evaluated on the \textit{warehouse\_huge}, \textit{warehouse\_huge2}, and \textit{warehouse\_huge3}. Layout of these huge maps are illustrated in \Cref{layhuge} in the appendix.

\subsubsection{Results}

The results of this group of experiments are shown in \Cref{wh_small,wh_large}. The x-axis displays the number of agents, and the y-axis shows the average throughput. \separ{} outperforms the other two learning-based methods on both maps. As the number of agents increased, \separ{} demonstrated better scalability compared to SCRIMP and PRIMAL2. 
When the number of agents is at most 128, learning-based methods perform comparably to LaCAM2, except that PRIMAL2 struggles at 128 agents. However, once the number of agents exceeds 128, \textbf{\separ{} still retains 50\% - 90\% of LaCAM2's performance, whereas the other two methods completely fail to cope, reaching only about 7\% - 15\% of \separ{}'s performance.}
Among all the methods, \separ{}-Follower fails to learn any effective routing mechanism, which proves the indispensability of imitation learning for maps with longer corridor lengths and higher effective edge sparsity.

From the runtime perspective, LaCAM2's average planning time per timestep grows superlinearly and worsens with the complexity of the map, while \separ{}'s average inference time scales linearly with the number of agents and remains unaffected by the complexity of the map. This implies that \separ{} demonstrates superior scalability on complex maps. 

\paragraph{Out of Distribution}

\begin{figure}[htbp]
    \centering
    \includegraphics[width=0.83\columnwidth]{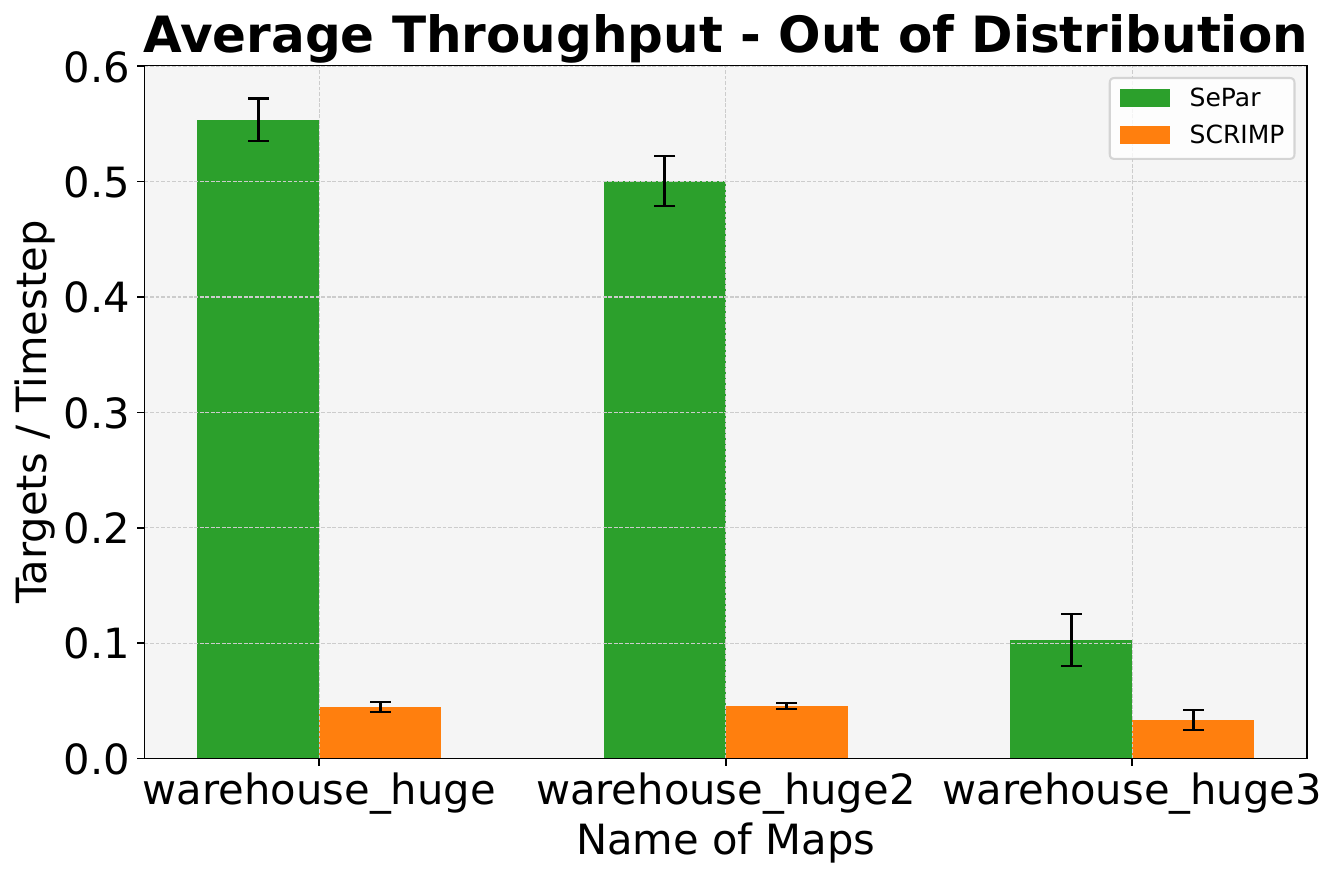}
    \caption{\separ{}'s performance on unseen maps, with the y-axis indicating the percentage of algorithm performance relative to LaCAM.} 
    \label{ood}
    \vspace{-15pt}
\end{figure}

The results of the experiments evaluating the generalizability of model are presented in \Cref{ood}. 
Since PRIMAL2 doesn't show any advantages in normal results, it is excluded. We use the planning results of LaCAM2 on these maps as an upper bound to normalize the evaluation results of the learning-based methods. Therefore, higher values indicate performance closer to that of LaCAM2. \separ{} significantly outperforms SCRIMP, owing to the robust representational capabilities of the Transformer. However, there remains a notable gap in performance when compared to the heuristic algorithm LaCAM2. Due to the significant differences in layout distribution between \textit{warehouse\_huge3} and the other maps, our method performs not well on \textit{warehouse\_huge3}.

\vspace{-5pt}

\subsection{One-shot and Lifelong MAPF Tasks in POGEMA}
\label{results in pogema}
\subsubsection{Experiment Setup}
To make a more extensive comparison with the well-known benchmarks, we compare \separ{} with various heuristic and learnable methods in POGEMA. For the MAPF tasks, we compare our method with LaCAM \cite{okumura2023lacam}, DCC, MAMBA \cite{egorov2022scalable}, and SCRIMP. And for LMAPF tasks, our main comparisons included MAMBA, Follower, Switcher (ASwitcher and Switcher) \cite{skrynnik2023switch}, and RHCR \cite{li2021lifelong}. Comprehensive comparative results, including other baselines, are presented in \Cref{full_mapf_results} in the appendix. Except for \separ{}, the rest of the data comes from files released by the authors of POGEMA. We train \separ{} on \textit{mazes} for both one-shot and lifelong MAPF scenarios. 
Consistent with POGEMA, we use 8 to 64 agents, with evaluation lengths of 128 for one-shot MAPF and 256 for lifelong MAPF.

\subsubsection{Results}
The results are shown in \Cref{mapf_results}. 
\Cref{mazes_csr,mazes_soc} respectively display the relationship between the success rate of MAPF tasks, the SoC, and the number of agents. 
In terms of success rate, \separ{} significantly outperforms other learnable methods.
Regarding the SoC, \separ{}'s advantage is more pronounced with a larger number of agents. 
With a smaller number of agents, \separ{} does not outperform some other learnable methods. 
\Cref{mazes_ath} shows the relationship between average throughput and the number of agents; it indicates that \separ{}'s performance is slightly inferior to that of Follower but surpasses all other methods except for RHCR and Follower.

Considering the results in \Cref{results in wh,results in pogema}, where the former involves tasks on maps with high PFCI values and the latter on maps with low PFCI values, our method not only demonstrates performance comparable to state-or-the-art algorithms on low-PFCI maps, but, more importantly, also shows clear advantages over other learning-based methods on highly structured, warehouse-like maps with high PFCI.

\section{Conclusion}
In this study, we introduce \separ{}, a multi-agent path finding method that combines MARL and imitation learning. 
\separ{} is a sequence modeling scheme that reduces decision complexity to a linear level while meeting the requirement for efficient information exchange between agents in dense, highly structured warehouse scenarios. 
Through extensive experiments, we demonstrate the advantages of \separ{} in MAPD tasks in warehouse scenarios and confirm the necessity of imitation learning. 
Additionally, results on POGEMA indicate that our method's advantages over other learnable methods are also significant. 
Future research will focus on improving our approach to handle random events in the environment, such as agent disconnections, sudden obstacles, and information disturbances in observations.

\vspace{-3pt}





\bibliographystyle{IEEEtran}
\bibliography{icra2026_reference}



\cleardoublepage
\appendices
\crefalias{section}{appendix}


\section{Proof Sketch of \Cref{Order-invariant}}
\label{app:proof_order}
By the chain rule, the conditional joint distribution $\pi(a_{1:n}\mid \bm{o})$ parameterized by $\theta$ under permutation $\sigma$ and $\nu$ can be written as repsectively:

\begin{equation}
    \begin{aligned}
    \pi^{\sigma}_{\theta}(a^{1:n}\mid \bm{o})=\prod_{k=1}^{n}\,\pi_{\theta}\big(a^{\sigma[k]}\mid \bm{o}, a^{\sigma[1]},\ldots,a^{\sigma[k-1]}\big),
    \end{aligned}
\end{equation}

\begin{equation}
    \begin{aligned}
    \pi^{\nu}_{\theta}(a^{1:n}\mid \bm{o})=\prod_{k=1}^{n}\,\pi_{\theta}\big(a^{\nu[k]}\mid \bm{o}, a^{\nu[1]},\ldots,a^{\nu[k-1]}\big).
    \end{aligned}
\end{equation}

Since both representations are just re-orderings of the same underlying joint distribution, the following equation holds:

\begin{equation}
    \begin{aligned}
    \pi_{\theta}(a^{1:n}\mid \bm{o}) 
    &= \prod_{k=1}^{n}\,\pi_{\theta}\big(a^{\sigma[k]}\mid \bm{o}, a^{\sigma[1]},\ldots,a^{\sigma[k-1]}\big) \\
    &= \prod_{k=1}^{n}\,\pi_{\theta}\big(a^{\nu[k]}\mid \bm{o}, a^{\nu[1]},\ldots,a^{\nu[k-1]}\big),
    \end{aligned}
\end{equation}

which implies that the joint distributions produced by $\pi^{\sigma}_{\theta}$ and $\pi^{\nu}_{\theta}$ are equivalent.

Since $f(\cdot)$ depends only on the induced autoregressive pathfinding policy $\pi$, the supremum of $f(\cdot)$ is independent of the decision order, which can be formalized as:

\begin{equation}
    \begin{aligned}
    \forall \sigma,\nu\in S_n\implies\sup_{\theta} f\big(\pi^{\sigma}_{\theta}\big)=\sup_{\theta} f\big(\pi^{\nu}_{\theta}\big).
    \end{aligned}
\end{equation}

This can be naturally extended to trajectory-level equivalence in Dec-POMDPs: 

\begin{equation}
    \begin{aligned}
    \forall \sigma,\nu\in S_n &\implies \\
    \sup_{\theta} \sum_{t=1}^T f\big(\pi^{\sigma}_{\theta}(a^{1:n}_t\mid \bm{o}_t)\big)&=\sup_{\theta} \sum_{t=1}^T f\big(\pi^{\nu}_{\theta}(a^{1:n}_t\mid \bm{o}_t)\big),
    \end{aligned}
\end{equation}

where $T$ is the length of trajectories, $a^{1:n;t}$ and $\bm{o}_t$ are the joint action and joint observation at timestep $t$.

\textbf{For lifelong MAPF}, let $\Delta^i_t=\mathbf{1}\{p^i_t=\tau^i_t\land \tau^i_{t-1}=\tau^i_t\land p^i_{t-1}\neq \tau^i_{t-1}\}$ denote whether agent $i$ first arrives its current goal $\tau^i_t$ at timestep $t$ and $p^i_t$ is the position of agent $i$ at timestep $t$. 

Then the number of goals completed by agents at timestep $t$ is:

\begin{equation}
    \begin{aligned}
    \label{traj-level}
    D_t=\sum_{i=1}^n\Delta^i_t,
    \end{aligned}
\end{equation}

which obviously depends only on the autoregressive pathfinding policy $\pi(\cdot\mid \bm{o}_t)$, besides the environment dynamics. Then we can use $D_{\pi,t}$ to denote $D_t$ under policy $\pi$.
By summing the above equation, we can obtain the number of goals completed by all agents up to timestep $T$ under policy $\pi$, that is to say, the throughput:

\begin{equation}
    \begin{aligned}
    N(\pi,T)=\sum_{t=0}^T D_{\pi,t},
    \end{aligned}
\end{equation}

From \Cref{traj-level}, the following holds:

\begin{equation}
    \begin{aligned}
    \label{eventual eq}
    \forall \sigma,\nu\in S_n\implies\sup_{\theta} N\big(\pi^{\sigma}_{\theta},T\big)=\sup_{\theta} N\big(\pi^{\nu}_{\theta},T\big).
    \end{aligned}
\end{equation}

\textbf{For oneshot MAPF}, we can simplify the definition of $\Delta^i_t$: 
\begin{equation}
    \begin{aligned}
    \Delta^i_t=\mathbf{1}\{p^i_t=\tau^i\}.
    \end{aligned}
\end{equation}

Then minimizing the makespan is equivalent to:

\begin{equation}
    \begin{aligned}
    T^\star = \min \left\{ T \in \mathbb{N}_{\ge 0} \;\middle|\; \max_{\pi} N(\pi,T) = n \right\},
    \end{aligned}
\end{equation}

which is essentially still maximizing $N(\pi,T)$, so \Cref{eventual eq} also applies to one-shot MAPF.

\section{Snapshot of the Warehouse Environment}

\begin{figure}[htbp]
\centering
\includegraphics[width=0.95\columnwidth]{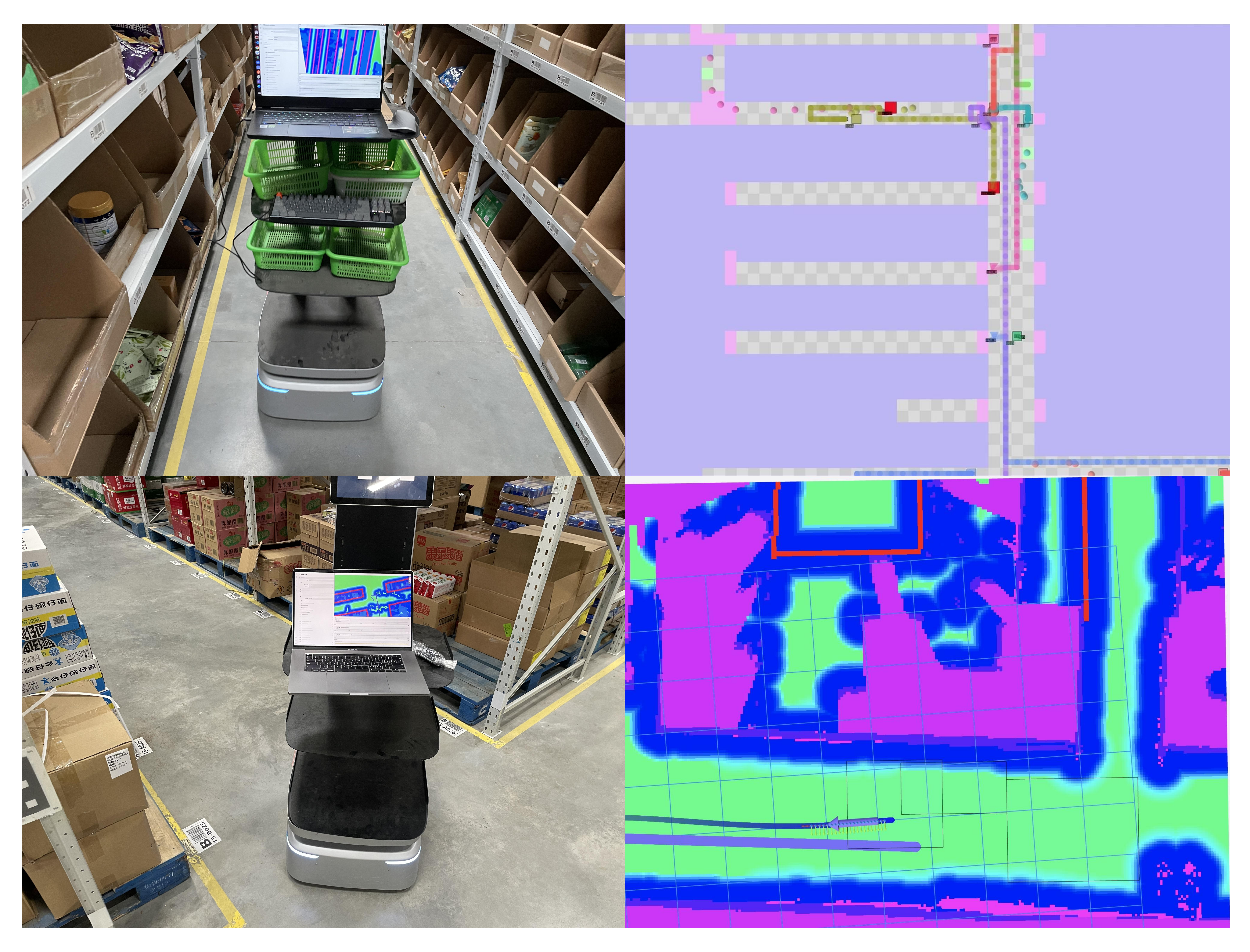} 
\caption{Snapshot of the physical and simulated shelves and robots deployed in the warehouse mockup. Left: Live scene images of shelves and robots in the warehouse. Right: Visualization in the simulated environments showing the obstacles, the robots and the trajectories.}
\label{snapshot}
\vspace{-15pt}
\end{figure}

Our simulation environment is an abstraction of real-world warehouses, as illustrated in \Cref{snapshot}.

\section{Detailed Data of PFCI}
\label{app:pfci}
This section presents the detailed data of Path Finding Complexity Index (PFCI), including maps from POGEMA and our warehouse simulator.

\section{Full results in the POGEMA}
\begin{figure*}[htbp]
    \centering
    \begin{subfigure}[t]{0.65\columnwidth}
        \includegraphics[width=\textwidth]{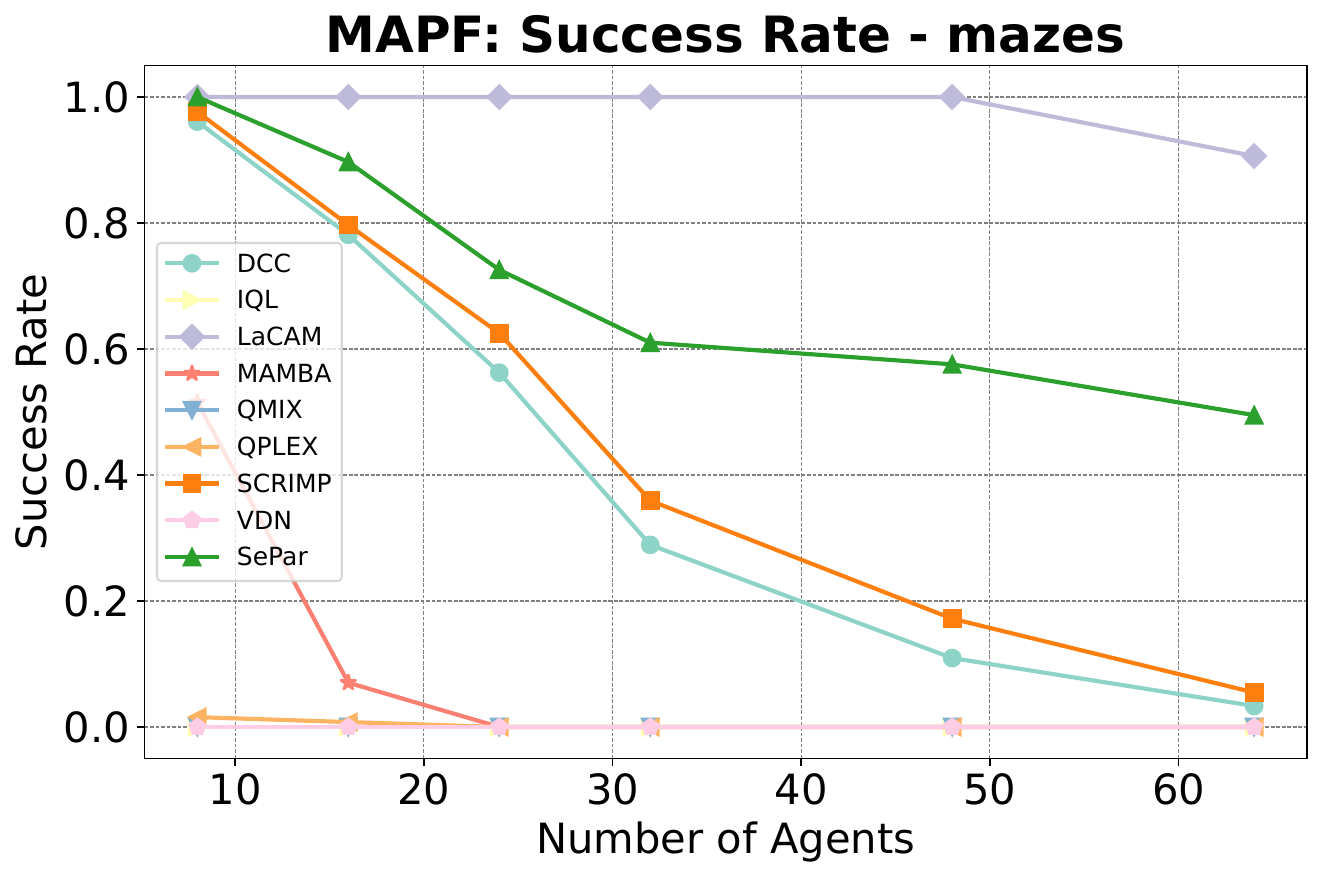} 
        \caption{}
        \label{full_mazes_csr}
    \end{subfigure}
    \hfill
    \begin{subfigure}[t]{0.664\columnwidth}
        \includegraphics[width=\textwidth]{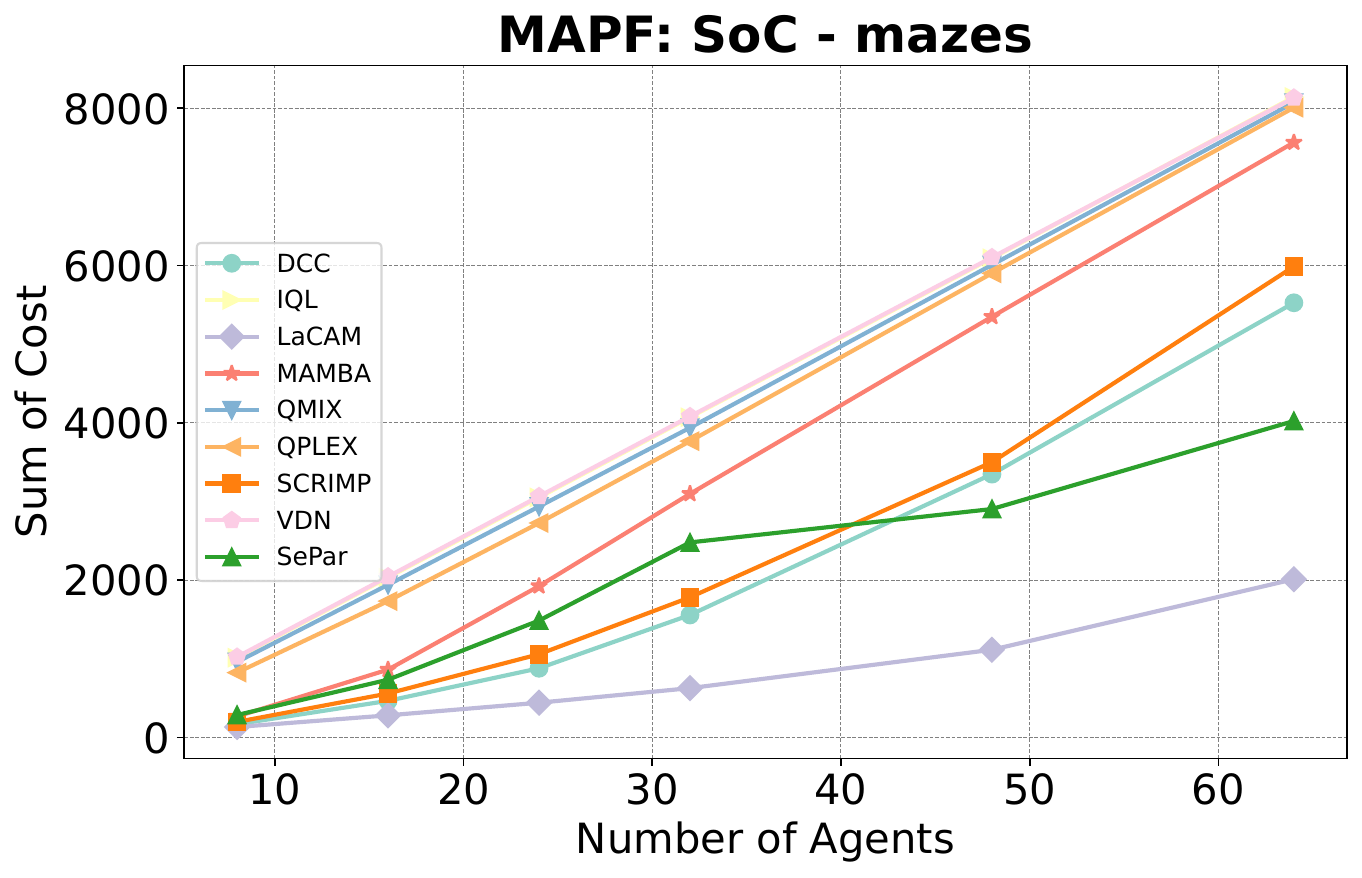} 
        \caption{}
        \label{full_mazes_soc}
    \end{subfigure}
    \hfill
    \begin{subfigure}[t]{0.65\columnwidth}
        \includegraphics[width=\textwidth]{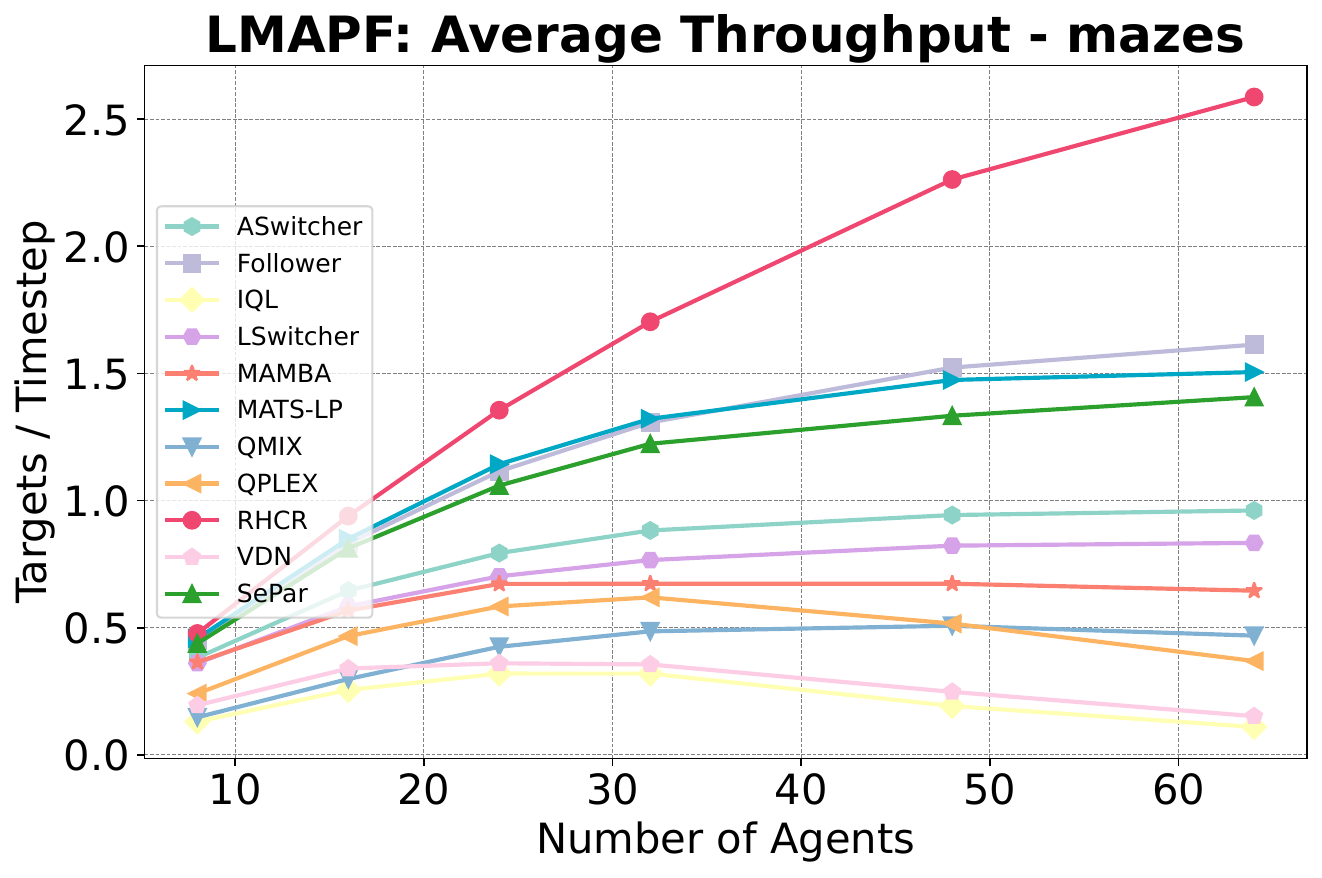}
        \caption{}
        \label{full_mazes_ath}
    \end{subfigure}
    \caption{Full Results in the POGEMA. Evaluation is conducted on \textit{mazes}. Figures \ref{full_mazes_csr} and \ref{full_mazes_soc} show the results of MAPF tasks, and Figure \ref{full_mazes_ath} shows the results of LMAPF tasks. In the MAPF tasks, \separ{} outperforms other learnable methods. In the LMAPF tasks, \separ{} ranks just behind Follower and RHCR, with the former being a state-of-the-art learnable solver for LMAPF and the latter representing a well-known planning solver.}
    \label{full_mapf_results}
    \vspace{-15pt}
    
\end{figure*}

\Cref{full_mapf_results} presents the results for all baselines involved in the POGEMA.

\section{Pseudo Code of Sequence Pathfinder}
\label{app:pse-code}
\begin{algorithm}[H]
\caption{Sequence Pathfinder (\separ{})}
\label{alg:separ}
\begin{algorithmic}[1]
\Require Stepsize $\alpha$, RL batch size $B_{\mathrm{RL}}$, IL batch size $B_{\mathrm{IL}}$, number of agents $n$, episodes $K$, steps per episode $T$, discount $\gamma$, FOV size $m$, A* prediction horizon $n_{\mathrm{pred}}$, heuristic algorithm $\mathcal{F}$
\State \textbf{Initialize:} encoder $\{\phi_0\}$, decoder $\{\theta_0\}$, replay buffer $\mathcal{B}_{\text{RL}}$, $\mathcal{B}_{\text{IL}}$ 
\For{$k=0,1,\dots,K-1$}
  \For{$t=0,1,\dots,T-1$}
    \State Collect observations $\bm{o}_t=\left( o^{i_1}_t,\dots,o^{i_n}_t \right)$ from environments.
    \Statex {\itshape // \textbf{Inference Phase}}
    \State Encode $\bm{o}_t$ to embeddings $\bm{z}_t=\left( z^{i_1}_t,\dots,z^{i_n}_t \right)$ with the observation feature extractor.
    \State Encode $\bm{z}_t$ to $\bm{\hat{o}}_t=\left( \hat{o}^{i_1}_t,\dots,\hat{o}^{i_n}_t \right)$ with the encoder.
    \State Feed $\bm{\hat{o}}_t$ to the decoder.
    \For{$m=0,1,\dots,n-1$}
      \State Given $a^{i_0}_t,\dots,a^{i_m}_t$ and $\bm{\hat{o}}_t$, infer $a^{i_{m+1}}_t$ with the auto-regressive decoder.
    \EndFor
    \State Execute joint actions $a^{i_0}_t,\dots,a^{i_n}_t$ and collect reward $R(\bm{o}_t,\bm{a}_t)$.
    \State Insert $(\bm{o}_t,\bm{a}_t,R(\bm{o}_t,\bm{a}_t))$ into $\mathcal{B}_{\text{RL}}$.
  \EndFor
  \Statex {\itshape // \textbf{Training Phase}}
  \Statex {\itshape // RL paradigm}
  \State Sample a random minibatch of $B_{\mathrm{RL}}$ steps from $\mathcal{B}_{\text{RL}}$.
  \State Use the encoder’s value head to get $V_{\phi}(\hat{o}^{i_1}),\dots,V_{\phi}(\hat{o}^{i_n})$.
  \State Compute $L_{\mathrm{Encoder}}(\phi)$ via \Cref{enc loss}.
  \State Compute the joint advantage $\hat{A}$ from $V_{\phi}(\hat{o}^{i_1}),\dots,V_{\phi}(\hat{o}^{i_n})$.
  \State Input $\hat{o}^{i_1},\dots,\hat{o}^{i_n}$ and $a^{i_0},\dots,a^{i_{n-1}}$; decode $\pi^{i_1}_{\theta},\dots,\pi^{i_n}_{\theta}$ in one pass.
  \State Compute $L_{\mathrm{Decoder}}(\theta)$ via \Cref{dec loss}.
  \State Update $\phi,\theta$ by minimizing $L_{\mathrm{Encoder}}(\phi)+L_{\mathrm{Decoder}}(\theta)$ with gradient descent.
  \Statex {\itshape // IL paradigm}
  \State Generate expert data with $\mathcal{F}$ on sampled states; collect $(\bm{o},\bm{a}^\star)$ and fill $\mathcal{B}_{\mathrm{IL}}$.
  \State Sample a random minibatch of $B_{\mathrm{IL}}$ steps from $\mathcal{B}_{\text{IL}}$.
  \State Input $\hat{o}^{i_1},\dots,\hat{o}^{i_n}$ and $a^{\star,{i_0}},\dots,a^{\star,{i_{n-1}}}$; decode $P(a^{*,i_1}_t|\pi^{i_1}_{\theta},\hat{o}^{i_1:i_n}_t;\phi,\theta), \dots, P(a^{*,i_n}_t|\pi^{i_n}_{\theta},\hat{o}^{i_1:i_n}_t;\phi,\theta)$ in one pass.
  \State Compute $L_{\mathrm{BC}}(\phi,\theta)$ via \Cref{bc loss}.
  \State Update $\phi,\theta$ by minimizing $L_{\mathrm{BC}}(\phi,\theta)$ with gradient descent.
\EndFor
\end{algorithmic}
\end{algorithm}

\begin{table}[htbp]
    \centering

    \caption{Metrics of Different Maps. The maps displayed above are from POGEMA, while those below are from our warehouse simulator. The complexity of the maps is positively correlated with the values of $\bm{v_e}$ and $\bm{l_{corr}}$.}
    \label{metrics}
        \resizebox{\linewidth}{!}{
    \begin{tabular}{c|c|c|c|c}\toprule
        \textit{Map} & \textit{$\rho_e$} & \textit{$\rho_t$} & \textit{$\bm{v_e}$}  & \textit{$\bm{l_{corr}}$} \\ \midrule 
        random & 0.0209 & 0.7969  & 0.0611 & 3.4801 \\ 
        mazes & 0.0200 & 0.7034  & 0.0737 & 4.4857 \\ 
        warehouse\_wfi & 0.0054 & 0.8419  & 0.2190 & 3.0000 \\ 
        movingai & 0.0026 & 0.7270  & 0.5440 & 3.4441 \\ 
        puzzles & 0.2827 & 0.6325  & 0.0059 & 5.4708 \\ \midrule
        warehouse\_small & 0.0038 & 0.3346  & 0.7777 & 14.6842 \\ 
        warehouse\_large & 0.0030 & 0.2060  & 1.6402 & 13.2436 \\ 
        warehouse\_huge & 0.0020 & 0.2233  & 2.2189 & 11.7480 \\ 
        warehouse\_huge2 & 0.0018 & 0.2478  & 2.2567 & 11.3565 \\ 
        warehouse\_huge3 & 0.0015 & 0.2920  & 2.2610 & 6.7557 \\ \bottomrule 
    \end{tabular}
    }
    \vspace{-15pt}
\end{table}

\section{Layout of Huge Warehouse Maps}
\label{huge}

\begin{figure*}[htbp]
\centering
\includegraphics[width=1.9\columnwidth]{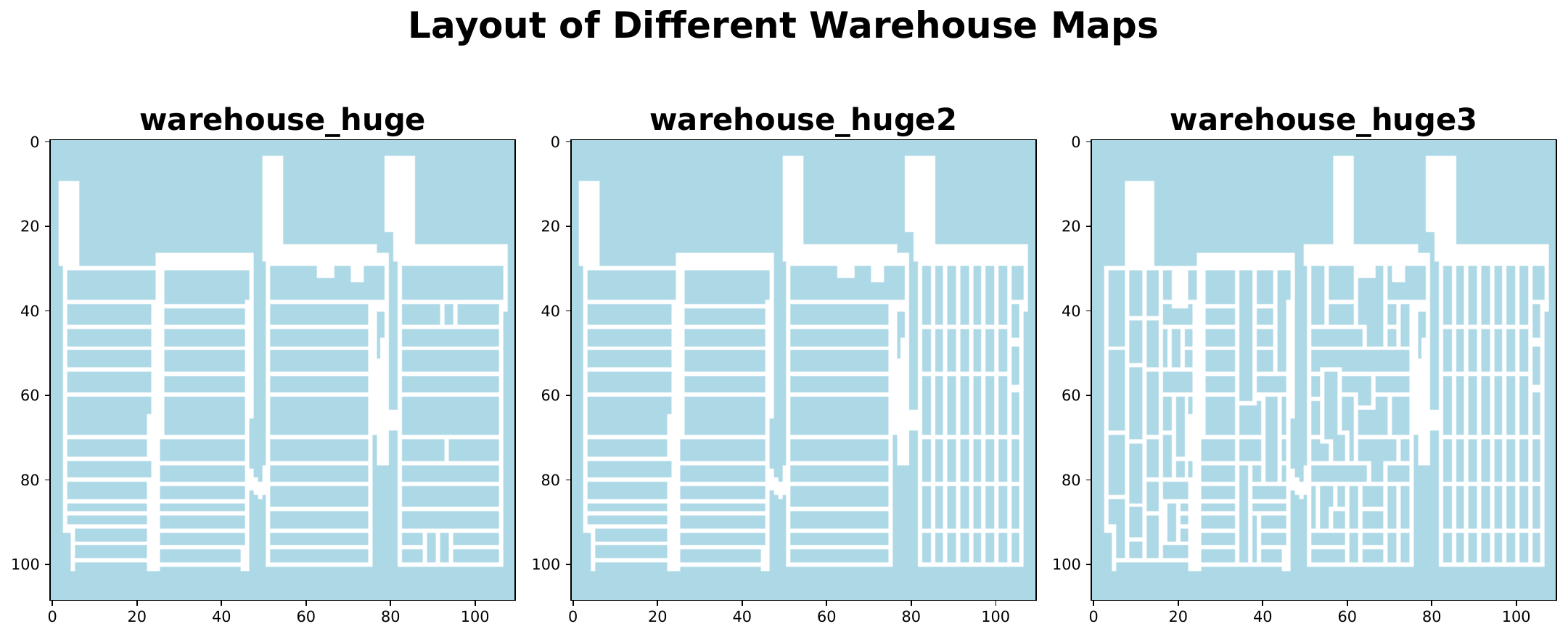}
\caption{Layout of different huge maps in the warehouse. The
traversable areas are represented in white, while obstacles
are indicated in light blue.}
\label{layhuge}
\vspace{-15pt}
\end{figure*}

\Cref{layhuge} presents the layout distributions of three large warehouse maps. Notably, the layout of \textit{warehouse\_huge3} differs significantly from those of the other two maps.

\end{document}